\documentclass[10pt,twocolumn,letterpaper]{article}

\usepackage{iccv}              
\usepackage{multirow} 
\usepackage{multirow} 

\definecolor{iccvblue}{rgb}{0.21,0.49,0.74}
\usepackage[pagebackref,breaklinks,colorlinks,allcolors=iccvblue]{hyperref}

\def\paperID{8181} 
\def\confName{ICCV}
\def\confYear{2025}

\title{CWNet: Causal Wavelet Network for Low-Light Image Enhancement}

\author{First Author\\
Institution1\\
Institution1 address\\
{\tt\small firstauthor@i1.org}
\and
Second Author\\
Institution2\\
First line of institution2 address\\
{\tt\small secondauthor@i2.org}
}

\author{
Tongshun Zhang$^{1,2}$ \quad 
Pingping Liu$^{1,2}$\thanks{Corresponding author} \quad 
Yubing Lu$^{1,2}$ \quad 
Mengen Cai$^{1,2}$ \quad 
Zijian Zhang$^{1,2}$ \\
Zhe Zhang$^{1,2}$ \quad 
Qiuzhan Zhou$^3$ \\[0.5ex]
$^1$College of Computer Science and Technology, Jilin University\\[0.5ex]
$^2$Key Laboratory of Symbolic Computation and Knowledge Engineering of Ministry of Education\\[0.5ex]
$^3$College of Communication Engineering, Jilin University\\[0.5ex]
{\tt\small \{tszhang23, luyb24, caime24, zhezhang23\}@mails.jlu.edu.cn,} \\
{\tt\small \{liupp, zhangzijian, zhouqz\}@jlu.edu.cn}
}

\begin{document}
\maketitle

\begin{abstract}
Traditional Low-Light Image Enhancement (LLIE) methods primarily focus on uniform brightness adjustment, often neglecting instance-level semantic information and the inherent characteristics of different features. To address these limitations, we propose CWNet (Causal Wavelet Network), a novel architecture that leverages wavelet transforms for causal reasoning. Specifically, our approach comprises two key components: 1) Inspired by the concept of intervention in causality, we adopt a causal reasoning perspective to reveal the underlying causal relationships in low-light enhancement. From a global perspective, we employ a metric learning strategy to ensure causal embeddings adhere to causal principles, separating them from non-causal confounding factors while focusing on the invariance of causal factors. At the local level, we introduce an instance-level CLIP semantic loss to precisely maintain causal factor consistency. 2) Based on our causal analysis, we present a wavelet transform-based backbone network that  effectively  optimizes the recovery of frequency information, ensuring precise enhancement tailored to the specific attributes of wavelet transforms. Extensive experiments demonstrate that CWNet significantly outperforms current state-of-the-art methods across multiple datasets, showcasing its robust performance across diverse scenes. Code is available at \href{https://github.com/bywlzts/CWNet-Causal-Wavelet-Network}{\textcolor{blue}{CWNet}}.
\end{abstract}

\section{Introduction}
\label{sec:intro}

LLIE is essential in computer vision, addressing challenges like dimness and detail loss that degrade image quality. While traditional methods like gamma correction \cite{moroney2000local}, Retinex theory \cite{ng2011total}, and histogram equalization \cite{pisano1998contrast} struggle with non-uniform lighting and extreme darkness, deep learning approaches \cite{cai2023retinexformer, zou2024wave, zhang2024dmfourllie, yang2024learning, yang2025learning, zhang2025cross} offer improved adaptability and performance. However, they often fail to fully exploit feature modeling and semantic information.

Frequency-based methods present a promising avenue for LLIE by effectively separating and enhancing high- and low-frequency information, improving detail and brightness while isolating noise. Nonetheless, existing methods \cite{wang2023fourllie, zhang2024dmfourllie, zou2024wave, li2023embedding, huang2022deep} treat frequency features uniformly, which limits their potential. Additionally, maintaining color and semantic consistency is a significant challenge, as many advanced methods \cite{jiang2023low, bai2024retinexmamba, zhang2024llemamba, wang2024uhdformer} often overlook these aspects, leading to visually unnatural or semantically inaccurate results. This paper addresses these gaps by exploring two key questions:

\textbf{\textit{Firstly, how can we ensure consistency in color and semantic information while improving lighting conditions?}} Current methods \cite{zhang2022deep,wu2023learning} often rely on color histogram-based losses to maintain color consistency, while SCL-LLE \cite{liang2022semantically} utilize downstream semantic segmentation consistency loss to enhance semantic brightness and color consistency. SKF \cite{wu2023learning} further improves semantic consistency at the feature level by extracting intermediate features through a semantic segmentation network. In contrast, the visual-language pre-trained model CLIP \cite{radford2021learning} demonstrates superior performance in maintaining color and semantic consistency. Many methods \cite{yang2023implicit, jiang2023low, zhang2024adapt} leverage CLIP to learn diverse features, achieving semantically guided enhancement. However, these approaches primarily focus on global semantic and color consistency, lacking the ability to ensure instance-level consistency.

\begin{figure}
    \centering
    \includegraphics[width=1.0\linewidth]{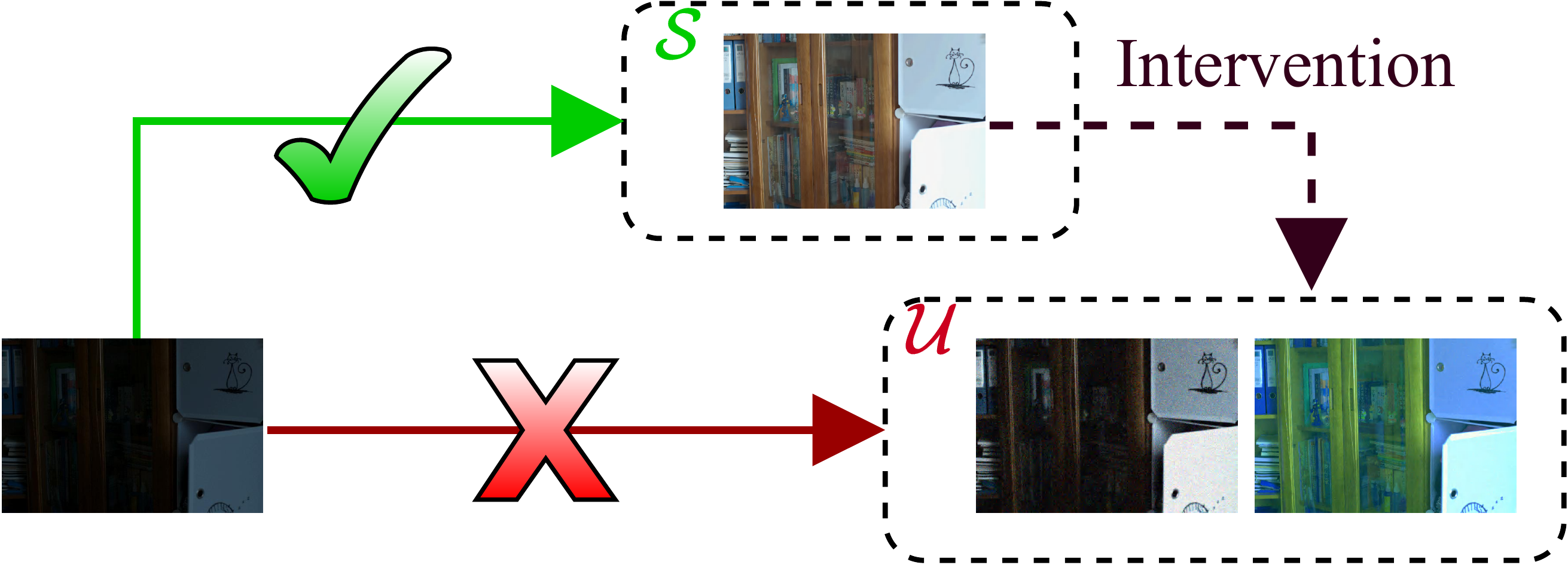}
    \vspace{-0.3cm}
    \caption{Structural causal model (SCM) for LLIE.}
    \vspace{-0.3cm}
    \label{fig:casual_ana}
\end{figure}

\textbf{\textit{Secondly, how can we establish a robust model that fully exploits frequency domain features?}} The two commonly used frequency domain transformations are Fourier transform and wavelet transform. Fourier-based methods \cite{huang2022deep, wang2023fourllie} excels in capturing global information by amplifying low-frequency components, enhancing overall brightness. However, its lack of spatial locality limits its ability to preserve high-frequency details like edges and textures, often resulting in brighter but less detailed images. Recent works\cite{li2023embedding, zhang2024dmfourllie} have improved detail preservation by incorporating phase processing, but challenges remain in achieving fine-grained detail enhancement and spatial coherence. In contrast, wavelet transforms provide superior spatial locality, effectively separating image content from noise and enhancing edge and texture details. However, wavelet-based methods~\cite{jiang2023low,zou2024wave} do not fully leverage the unique characteristics of the frequency domain, which limits their recovery potential.

To address these limitations, we propose a Structural Causal Model (SCM)~\cite{scholkopf2022causality} for Low-Light Image Enhancement (LLIE) based on causal inference, as shown in Fig.~\ref{fig:casual_ana}. Within this framework, we establish the core objective of the LLIE task: to maintain consistency in causal factors $\mathcal{S}$ (semantic information). Meanwhile, non-causal factors $\mathcal{U}$ (color and brightness anomalies) need to be filtered out. This causal perspective enables us to effectively distinguish meaningful semantic content from confounding degradations. Building on this causal analysis, we propose a two-level strategy. At the global level, we obtain non-causal factors $\mathcal{U}$ through intervention procedures. Subsequently, we employ a causally-guided metric learning approach to filter out non-causal factors in the latent space. At the local level, we introduce an instance-level CLIP semantic loss to maintain fine-grained semantic consistency for each instance, achieving the objective of ensuring that causal factors $\mathcal{S}$ remain consistent.

Based on the SCM, we meticulously propose a wavelet-based backbone network to support this causal concept, referred to as the Causal Wavelet Network (CWNet). CWNet incorporates a Hierarchical Feature Restoration Block (HFRB) after each sampling layer, consisting of three components: a Feature Extraction (FE), a High-Frequency Enhancement Block (HFEB), and a Low-Frequency Enhancement Block (LFEB). The FE adaptively extracts wavelet frequency domain features and compensates for missing information through interaction. HFEB, inspired by State Space Models (SSM) \cite{zou2024wave, guo2025mambair, zhang2024llemamba}, employs a 2D Selective Scanning Module (2D-SSM) aligned with the scanning order of wavelet high-frequency components, enabling accurate recovery of high-frequency details. For low-frequency information, we develop the LFEB module for comprehensive recovery.

In summary, our main contributions are as follows:  
\begin{itemize}  
    \item We introduce a novel causal framework for LLIE, separating causal and non-causal factors to enhance image quality while preserving semantics.
    \item We propose a two-level consistency strategy, combining causally-guided metric learning for global consistency and instance-level CLIP loss for local semantic and color consistency. 
    \item We develop the Causal Wavelet Network (CWNet) with a Hierarchical Feature Restoration Block (HFRB) to model wavelet frequency features, enabling precise high-frequency recovery and robust low-frequency handling. 
    \item Extensive experiments validate CWNet’s state-of-the-art performance across diverse datasets, demonstrating its robustness and scalability.
\end{itemize}  

\section{Related Work}
\label{sec:relatedwork}

\textbf{Low-Light Image Enhancement (LLIE):}
LLIE methods can be categorized into non-learning-based and learning-based approaches. Traditional techniques, such as histogram equalization (HE) \cite{pisano1998contrast} and Retinex theory \cite{ng2011total}, enhance images by improving contrast or adjusting illumination and reflectance maps. With the advent of deep learning, methods like LLNet \cite{lore2017llnet} and Deep Retinex Decomposition \cite{wei1808deep} combined Retinex theory with neural networks, leading to significant advancements. Recent approaches, such as FourLLIE \cite{wang2023fourllie} and DMFourLLIE \cite{zhang2024dmfourllie}, leverage frequency domain features to enhance brightness, while Retinexformer \cite{cai2023retinexformer} integrates transformers to address long-range dependencies. Advanced frameworks like UHDformer \cite{wang2024uhdformer} and LightDiff \cite{li2024light} tackle ultra-high-definition restoration and unpaired low-light enhancement, respectively, expanding LLIE's scope to more complex tasks.

\textbf{State Space Models (SSM):}
State Space Models (SSMs) have emerged as efficient alternatives to CNNs and Transformers for handling long-range dependencies, with linear scalability \cite{gu2021efficiently}. Mamba, a structured SSM, has been applied to tasks like super-resolution \cite{huang2024irsrmamba}, image classification \cite{shi2024vmambair}, and restoration \cite{guo2025mambair}. In LLIE, Retinexmamba \cite{bai2024retinexmamba} integrates SSMs into Retinexformer for faster processing, while Wave-Mamba \cite{zou2024wave} explores UHD low-light enhancement. Recent innovations include LocalMamba \cite{huang2024localmamba}, which uses localized scanning for detail preservation, and LLEMamba \cite{zhang2024llemamba}, which employs bidirectional scanning to balance local and global focus. These advancements highlight the growing role of SSMs in LLIE.

\textbf{Causal Inference:}
Causal inference focuses on identifying and quantifying causal relationships, with growing applications in computer vision. CIIM \cite{yan2024causality} removes modality bias through causal intervention, while DCIN \cite{li2023towards} applies causal reasoning to reduce knowledge bias in image-text matching. MuCR \cite{li2024multimodal} embeds semantic causal relationships for image synthesis. Despite its potential, causal inference remains underexplored in low-level image processing tasks, presenting an opportunity for innovation in LLIE.

\section{Method}
\label{method}

\begin{figure}
    \centering
    \vspace{-0.3cm}
    \includegraphics[width=1.0\linewidth]{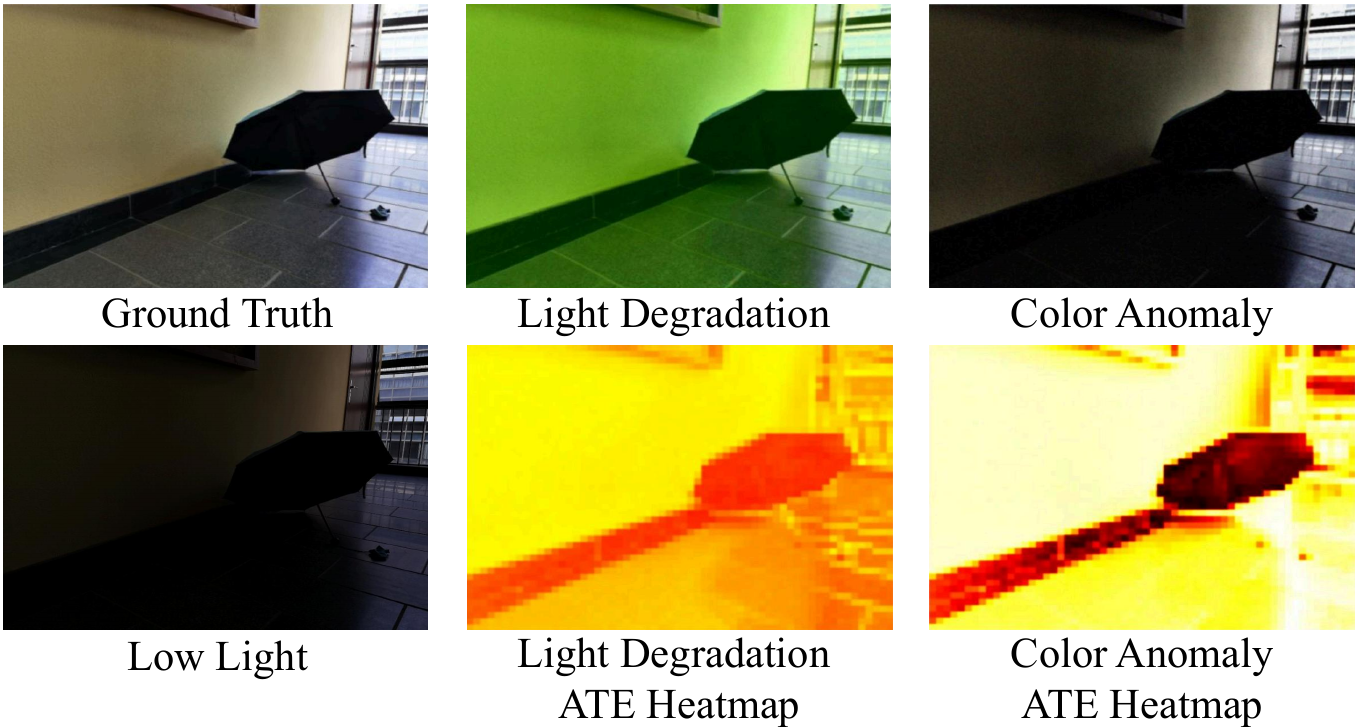}
    \caption{ATE Heatmap Analysis (PSNR). Top row: Ground truth, light degradation, and color anomaly examples. Bottom row: Low light input, ATE heatmaps for light degradation and color anomaly. Brighter regions indicate greater sensitivity to degradations.}
    \vspace{-0.3cm}
    \label{fig:ate_heatmap}
\end{figure}

\subsection{Causal Inference Analysis for LLIE}

\subsubsection{Meaningful and Harmless Causal Interventions}  
\label{sec:casual_interventions}  
For effective causal analysis, interventions must be meaningful and harmless. To achieve these aims, we refer to \cite{guo2024onerestore} and design two types of interventions applied to ground truth (normal-light) images for synthetic degradation:

\textbf{Light Degradation Intervention}: Instead of simple ablation (which violates the harmless principle), we utilize a physics-based illumination degradation model. Given a normal-light image $I$, we generate a light-degraded version $I_{l}$ as follows:

\begin{equation}  
I_{l} = \frac{I}{L}L^\gamma + \varepsilon,  
\end{equation}  

where $L$ is the illumination map generated through LIME \cite{lime}. Here, $\gamma \in [2, 5]$ controls the severity of the degradation, and $\varepsilon$ is Gaussian noise with mean 0 and variance in $[0.03, 0.08]$. This approach offers substantial yet realistic lighting changes while preserving the original semantic content.
    
\textbf{Color Anomaly Intervention}: To assess the impact of color distortion, we apply the following transformation to the ground truth image:

\begin{equation}  
I_{c} = \Delta H(I) + \Delta S(I) + \sum_{K=R,G,B} \Delta K(I) + \varepsilon,  
\end{equation}  

where $\Delta H \in [-30, 30]$ represents hue shift, $\Delta S \in [-50, 50]$ represents saturation shift, and $\Delta K \in [-50, 50]$ denotes RGB channel offsets.

\begin{figure}[t]
    \centering  
    \begin{subfigure}[t]{\linewidth}  
        \centering  
        \includegraphics[width=1.0\linewidth]{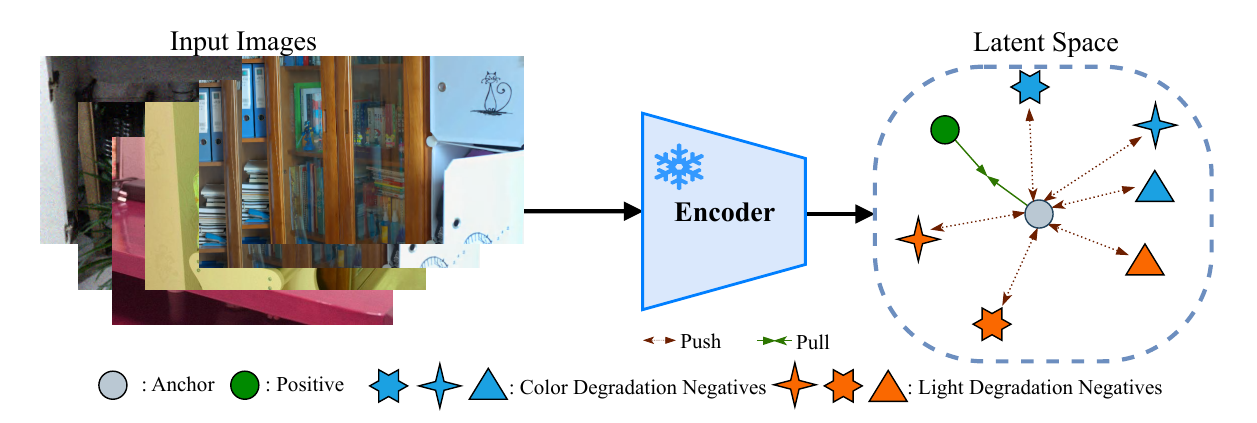}
        \caption{Causality-Driven Metric Learning Strategy for Causal Inference.The latent space organization includes: Anchor (gray): The network-processed low-light image. Positive (green): The ground truth normal-light reference sharing the same semantic causal factors as the anchor. Color Degradation Negatives (blue): Counterfactual samples generated through color perturbation interventions on normal-light images from different scenes. Light Degradation Negatives (orange): Counterfactual samples generated through brightness perturbation interventions on normal-light images from different scenes.}  
    \end{subfigure}  
    \begin{subfigure}[t]{\linewidth}  
        \centering  
        \includegraphics[width=1.0\linewidth]{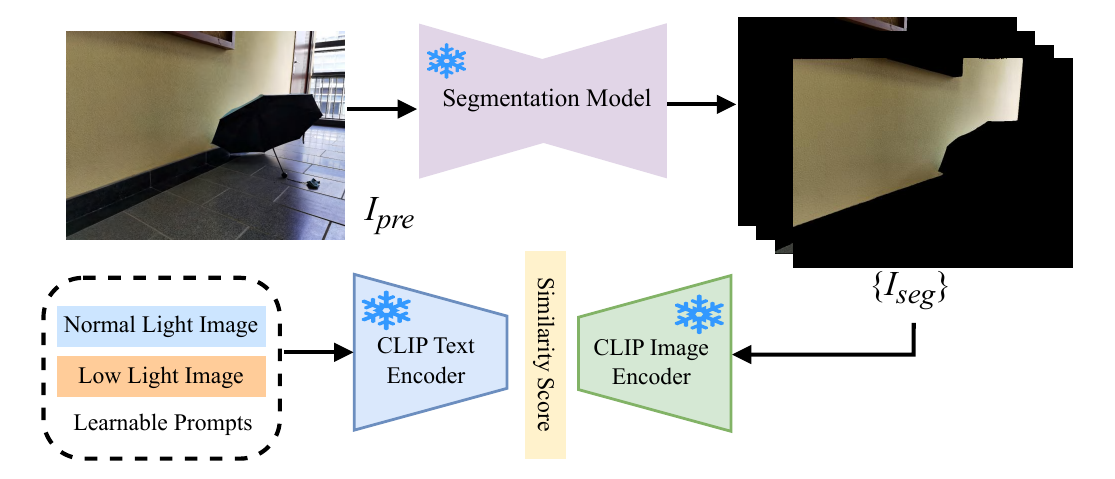}
        \caption{Instance-Level CLIP Semantic Loss. The enhanced image result $I_{pre}$
  is processed through a pre-trained segmentation network to obtain a series of segmented instance sub-images ${I_{seg}}$. Each sub-image is then iteratively aligned with corresponding textual prompts to ensure semantic consistency.}  
    \end{subfigure}  
    \caption{Global and local causal intervention methods. (a) Eliminate global non-causal interference from illumination and color. (b) Ensure causal semantic consistency.}  
    \label{fig:cas_operation}  
\end{figure} 

\begin{figure*}
    \centering
    \includegraphics[width=1\linewidth]{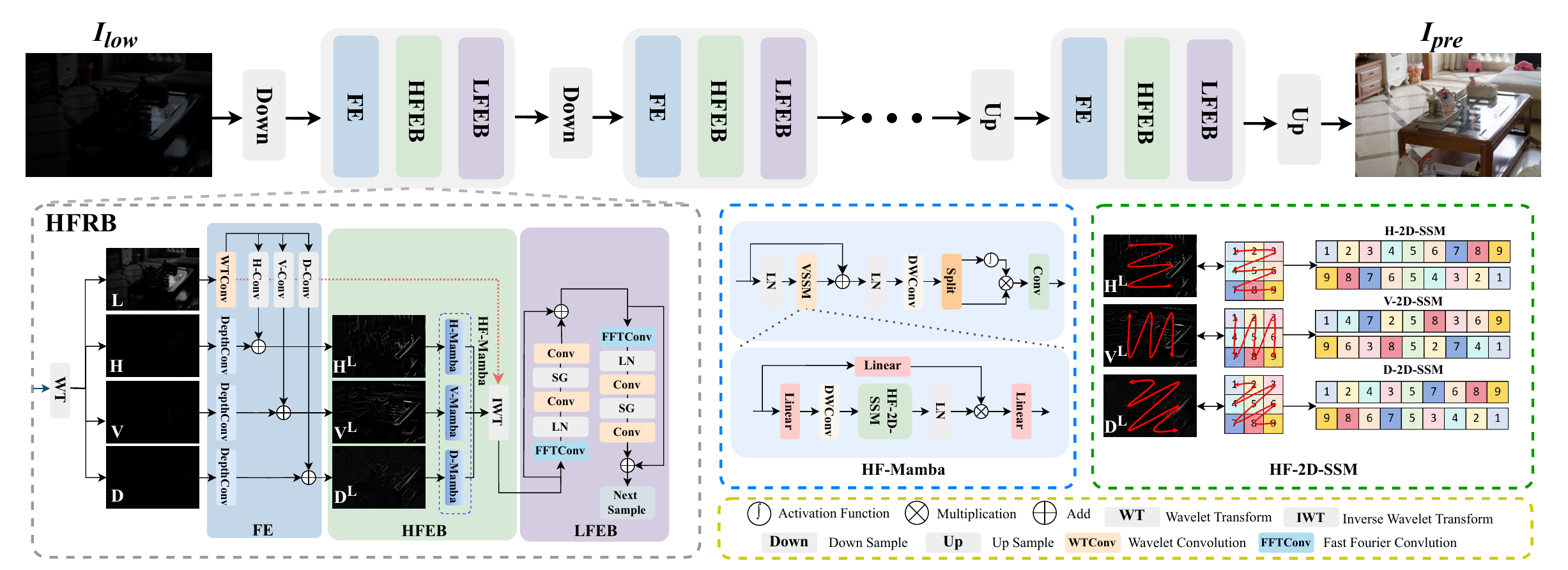}
    \vspace{-0.4cm}
    \caption{Overall Architecture of CWNet. The low-light image $I_{low}$ is processed through the sampling and HFRB modules to generate the predicted image $I_{pre}$. The expanded structures of each module are shown below.}
    \vspace{-0.4cm}
    \label{fig:network}
\end{figure*}

\subsubsection{Average Treatment Effect Analysis}  
\label{sec:ate_analysis}  

To quantitatively assess the impact of our interventions on different image regions, we employ Average Treatment Effect (ATE) analysis~\cite{angrist1995identification, vanderweele2013causal}. 

For a feature or region \( p_i \), the ATE is calculated as:
\begin{equation}  
\phi_{\mathcal{F}}[p_i] = \mathbb{E}\{\mathcal{M}_R(I)\} - \mathbb{E}_{t\in\{1:T\}}\{\mathcal{M}_R(I|do(p_i = x_t))\} ,
\end{equation}  

where \( I \) is the ground truth image, \( \mathcal{M}_R \) represents the quality metric (PSNR), and \( I|do(p_i = x_t) \) denotes the image with our interventions (light degradation, color anomaly, or noise) applied at intensity \( x_t \).

To visualize region-specific sensitivity, we compute:
\begin{equation}  
\phi_{\mathcal{F}}(I) = \{\phi_{\mathcal{F}}[p_i]\}_{i=1}^N,
\end{equation}  

creating attribution maps that highlight regions most affected by the interventions. As illustrated in Fig.~\ref{fig:ate_heatmap}, this analysis reveals that degradations affect different semantic regions with varying intensities, emphasizing the need for both global causal consistency and instance-level causal preservation in our enhancement approach.

\subsubsection{Causally-Guided Metric Learning}  
\label{sec:casual_metric}

We first address global causal consistency through a causally-guided metric learning approach within a causal inference framework. As shown in Fig.~\ref{fig:cas_operation}(a), we introduce a sample mining strategy for metric learning to disentangle illumination-invariant semantic features (causal factors) from degradation-related factors (non-causal factors). The processed low-light image serves as the anchor, paired with its corresponding normal-light reference as the positive sample. The low-light image, as an extremely degraded instance of the reference in brightness and color (non-causal factors), forms an intrinsic hard positive pair, compelling the metric learning to focus on semantic invariance. Furthermore, we construct counterfactual negative samples by perturbing color and brightness in normal-light images from different scenes. This strategy deliberately excludes other low-light images (preventing confusion between causal and non-causal features), instead forcing the model to discriminate fundamental semantic differences (divergent causal factors) even when non-causal features may resemble the anchor. The metric loss is defined as:  
\vspace{-0.2cm}
\begin{equation}
\resizebox{0.45\textwidth}{!}{$
\mathcal{L}_{\mathrm{ca}}(F_{p}, \hat{F},\{F_{l}\}\,\{F_{c}\})=\frac{\mathcal{L}_{1}(F_{p},\hat{F})}{\xi(\sum_{l=1}^{L}\mathcal{L}_{1}(F_{l},\hat{F}) +\sum_{c=1}^{C}\mathcal{L}_{1}(F_{c},\hat{F}))} 
$,}
\end{equation} 

where \( F_{p}, \hat{F}, \{F_{l}\}, \{F_{c}\} \) are the feature representations of the positive sample, anchor, light negative samples, and color abnormal negative samples, respectively. The hyperparameter \( \xi = \frac{1}{L+C} \) normalizes the contributions from light and color negative samples, with \( L \) and \( C \) denoting their total counts.  

\subsubsection{Instance-Level Causal Consistency through CLIP}  
\label{sec:clip_causal}

While our metric learning approach ensures global causal consistency, our ATE analysis identified significant region-specific variability in degradation sensitivity. To uphold local semantic integrity, we introduce an instance-level CLIP-based causal consistency module.

As depicted in Fig.~\ref{fig:cas_operation}(b), we employ HRNet \cite{wang2020deep}, pre-trained on PASCAL-Context \cite{mottaghi2014role}, to extract semantic instance maps. These maps, along with text prompts, are processed by CLIP encoders to assess semantic consistency:

\begin{equation}  
\hat{y} = \frac{1}{K}\sum_{k=1}^{K}  
\frac{e^{\cos(\Phi_{\mathrm{image}}(I_{seg}^k), \Phi_{\mathrm{text}}(T_{low}))}}  
{\sum_{i\in\{low, normal\}} e^{\cos(\Phi_{\mathrm{image}}(I_{seg}^k), \Phi_{\mathrm{text}}(T_{i}))}},
\end{equation}  

where \( K \) is the number of sub-instance maps and \( I_{seg}^k \) is a sub-instance map. We optimize using cross-entropy loss:

\begin{equation}  
L_{sem} = -\left( y \log(\hat{y}) + (1 - y) \log(1 - \hat{y}) \right),
\end{equation}  

where $y$ is the label of the current image, $0$is for low light image and $1$ is for normal light image.

\subsection{Causal Wavelet Network (CWNet)}

The architecture of CWNet is illustrated in Fig.\ref{fig:network}, comprising upsampling and downsampling layers along with the HFRB. The HFRB consists of FE, HFEB, and LFEB. Below is a detailed introduction to each component:

\subsubsection{Feature Extraction (FE)}

Given a low-light image \(I_{low}\in\mathbb{R}^{H \times W \times C}\), we use wavelet transform (WT) to decompose it into four different frequency sub-bands:  
\vspace{-0.1cm}
\begin{equation}
\resizebox{0.45\hsize}{!}{$\{L,H,V,D\}=WT(I_{low})$,}
\end{equation}
where $L,H,V,D\in\mathbb{R}^{\frac H2\times\frac W2\times C}$ represent low-frequency component,  horizontal, vertical and diagonal high-frequency components, respectively. The image can be reconstructed from these frequency sub-bands using inverse wavelet transform (IWT):
\vspace{-0.1cm}
\begin{equation}
\resizebox{0.45\hsize}{!}{$I_{low} = IWT\{L,H,V,D\}$.}
\end{equation}

In the FE, we design the extraction process to leverage the wavelet domain's characteristics. High-frequency features are obtained using depthwise separable convolutions, while low-frequency features utilize WTConv \cite{finder2025wavelet}. The WTConv layer achieves a larger receptive field without additional parameter complexity, essential for capturing low-frequency information. Studies \cite{jiang2023low, zou2024wave} indicate that most information resides in low-frequency components, while high-frequency details are less sensitive in low-light scenarios. To enhance extraction, we employ three convolution kernels aligned with high-frequency directions. The extraction process is formalized as:
\begin{equation}
\resizebox{0.3\textwidth}{!}{$
\begin{gathered}
L^{\prime} = WTConv(L), \\
H^{L} = DepthConv(H) + H\text{-}Conv(L^{\prime}),\\
V^{L} = DepthConv(V) + V\text{-}Conv(L^{\prime}),\\
D^{L} = DepthConv(D) + D\text{-}Conv(L^{\prime}),
\end{gathered} $}
\end{equation}
where \(L^{\prime}, H^{L}, V^{L}, D^{L}\) denote the features extracted post-FE, with \(H\text{-}Conv\), \(V\text{-}Conv\), and \(D\text{-}Conv\) extracting horizontal, vertical, and diagonal features, respectively. For detailed structures, please refer to our supplementary materials.

\subsubsection{High-Frequency Enhancement Block (HFEB)}  
Recent advancements in state space models (SSM) have significantly improved image enhancement tasks \cite{bai2024retinexmamba, huang2024irsrmamba, zou2024wave, zhang2024llemamba}. 

SSMs transform one-dimensional signals into outputs via latent state representations through linear ordinary differential equations. For a system with input \( x(t) \) and output \( y(t) \), the model dynamics are described by:
\begin{equation}
\begin{gathered}
h^{\prime}(t) = A h(t) + B x(t),  \quad
y(t) = C h(t) + D x(t),
\end{gathered}
\end{equation}
where \( A \), \( B \), \( C \), and \( D \) are system parameters. The discrete versions, such as Mamba, utilize the zero-order hold (ZOH) discretization, represented as follows:
\begin{equation}
\begin{gathered}
h_{t}^{\prime} = \bar{A} h_{t-1} + \bar{B} x_{t},  \quad
y_{t} = C h + D x_{t}, \\
\bar{A} = e^{\Delta A},  \quad
\bar{B} = (\Delta A)^{-1}(e^{\Delta A} - I) \cdot \Delta B,
\end{gathered} 
\end{equation}
where \( \bar{A} \) and \( \bar{B} \) are the discrete counterparts of \( A \) and \( B \).

Inspired by SSM, we propose the High-Frequency Mamba (HF-Mamba) module, designed for wavelet high-frequency components. The HF-Mamba consists of three parts: D-Mamba, V-Mamba, and H-Mamba. As illustrated in Fig.\ref{fig:network}, it applies Layer Normalization (LN), followed by a Visual State Space Module (VSSM) with residual connections, and culminates in a gated feedforward network to enhance channel information flow.

While many existing methods \cite{zou2024wave, guo2025mambair, bai2024retinexmamba, huang2024irsrmamba, zhang2024llemamba} directly adapt the 2D-SSM structure from VMamba \cite{shi2024vmambair}, we propose distinct processing for wavelet high-frequency components. Horizontal scanning utilizes H-2D-SSM, vertical scanning employs V-2D-SSM, and diagonal scanning incorporates D-2D-SSM. This approach is formalized as:
\begin{equation}
\resizebox{0.19\textwidth}{!}{$
\begin{gathered}
\tilde{H}^{L} = H\text{-}2D\text{-}SSM(H^{L}), \\
\tilde{V}^{L} = V\text{-}2D\text{-}SSM(V^{L}), \\
\tilde{D}^{L} = D\text{-}2D\text{-}SSM(D^{L}).
\end{gathered} $}
\end{equation}
This consistent scanning of high-frequency features extracted by the FE further enhances detail representation.

\subsubsection{Low-Frequency Enhancement Block (LFEB)}
Previous works \cite{zhu2020eemefn, dong2022incremental, xu2023low} have employed structure-guided enhancement techniques to optimize image generation, demonstrating that refined high-frequency components can significantly aid in generation and restoration tasks, particularly in LLIE. Building on these insights, we enhance high-frequency components post-High-Frequency Enhancement Block (HFEB) and reconstruct the frequency domain using Inverse Wavelet Transform (IWT):

\begin{equation}
\resizebox{0.55\hsize}{!}{$\bar{I}_{low} = IWT\{L^{\prime},\tilde{H}^{L},\tilde{V}^{L},\tilde{D}^{L}\}$,}
\end{equation}

where the reconstructed image \( \bar{I}_{low} \) serves as input to the LFEB. Fast Fourier Convolution (FFC) \cite{chi2020fast} integrates global context within early neural network layers using channel Fast Fourier Transform (FFT) for a broader receptive field.

Based on the above, we propose an LFEB, illustrated in Fig. \ref{fig:network}, consisting of two residual blocks tailored for processing low-frequency components, which require large receptive fields. Both blocks employ Fast Fourier Convolution to enhance global features.
The fist block apply a 5×5 convolution with appropriate padding to expand the receptive field for local spatial context, while the SimpleGate mechanism ensures efficient information flow with minimal loss during activation. Finally, a 1×1 convolution restores the feature dimensions. The second block emphasizes inter-channel correlations and feature enhancement.  It employs channel expansion through a 1×1 convolution that quadruples the channel dimensions.The SimpleGate mechanism selectively preserves important features, and another 1×1 convolution compresses the features back to the original channel size.
Following the LFEB and subsequent modules, low-frequency components are refined under the guidance of high-frequency components, resulting in the predicted brightened output \( I_{pre} \).

\begin{table*}[ht]  
\centering  
\resizebox{1.0\textwidth}{!}{  
\begin{tabular}{l|l|ccc|ccc|ccc|ccc|c|c}  
\toprule  
\multirow{2}{*}{\textbf{Category}} & \multirow{2}{*}{\textbf{Methods}} & \multicolumn{3}{c|}{\textbf{LOL-v1}} & \multicolumn{3}{c|}{\textbf{LOL-v2-Real}} & \multicolumn{3}{c|}{\textbf{LOL-v2-Syn}} & \multicolumn{3}{c|}{\textbf{LSRW-Huawei}} & \multirow{2}{*}{\textbf{\#Param}} & \multirow{2}{*}{\textbf{\#Flops}} \\   
\cmidrule(lr){3-5} \cmidrule(lr){6-8} \cmidrule(lr){9-11} \cmidrule(lr){12-14}  
& & PSNR ↑ & SSIM ↑ & LPIPS ↓ & PSNR ↑ & SSIM ↑ & LPIPS ↓ & PSNR ↑ & SSIM ↑ & LPIPS ↓ & PSNR ↑ & SSIM ↑ & LPIPS ↓ & \textbf{(M)} & \textbf{(G)} \\   
\midrule  
\multirow{3}{*}{Traditional} & NPE~\cite{npe} & 16.97 & 0.5928 & 0.2456 & 17.33 & 0.4642 & 0.2359 & 16.60 & 0.7781 & 0.1079 & 17.08 & 0.3905 & 0.2303 & - & - \\   
& LIME~\cite{lime} & 16.76 & 0.4440 & 0.2060 & 15.24 & 0.4190 & 0.2203 & 16.88 & 0.7578 & 0.1041 & 17.00 & 0.3816 & 0.2069 & - & - \\   
& SRIE~\cite{retinex1} & 11.80 & 0.5000 & 0.1862 & 14.45 & 0.5240 & 0.2160 & 14.50 & 0.6640 & 0.1484 & 13.42 & 0.4282 & 0.2166 & - & - \\   
\midrule
\multirow{3}{*}{CNN-based} & Kind~\cite{kind} & 20.87 & 0.7995 & 0.2071 & 17.54 & 0.6695 & 0.3753 & 22.62 & 0.9041 & 0.0515 & 16.58 & 0.5690 & 0.2259 & 8.02 & 34.99 \\   
& MIRNet~\cite{lowlight9} & 24.14 & 0.8305 & 0.2502 & 22.11 & 0.7942 & 0.1448 & 22.52 & 0.8997 & 0.0568 & 19.98 & 0.6085 & 0.2154 & 31.79 & 785.1 \\   
& Kind++~\cite{kind++} & 17.97 & 0.8042 & 0.1756 & 19.08 & 0.8176 & 0.1803 & 21.17 & 0.8814 & 0.0678 & 15.43 & 0.5695 & 0.2366 & 8.27 & 2970.5 \\    
 
\midrule
\multirow{3}{*}{Frequency-based} & FourLLIE~\cite{wang2023fourllie} & 20.99 & 0.8071 & 0.0952 & 23.45 & 0.8450 & 0.0613 & 24.65 & 0.9192 & 0.0389 & 21.11 & 0.6256 & 0.1825 & 0.12 & 4.07 \\   
& UHDFour~\cite{li2023embedding} & 22.89 & 0.8147 & 0.0934 & 27.27 & 0.8579 & 0.0617 & 23.64 & 0.8998 & 0.0341 & 19.39 & 0.6006 & 0.2466 & 17.54 & 4.78 \\   
& DMFourLLIE \cite{zhang2024dmfourllie} & 22.98&  0.8273&  0.0792&  26.40 & 0.8765&  0.0526 & 25.74 & 0.9308 & \underline{0.0251} & 21.09 & \underline{0.6328} & 0.1804 & 0.41 & 1.70 \\  
\midrule
\multirow{3}{*}{Transformer-based} 
& SNR-Aware~\cite{lowlight8} & \textbf{23.93} & \underline{0.8460}& 0.0813& 21.48 &0.8478& 0.0740 & 24.13 & 0.9269 & 0.0318 & 20.67 & 0.5911 & 0.1923 & 39.12 & 26.35 \\  
& Retinexformer~\cite{cai2023retinexformer} & 22.71& 0.8177& 0.0922 &24.55 &0.8434 &0.0627 & 25.67 & 0.9295 & 0.0273 & \underline{21.23} & 0.6309 & 0.1699 & 1.61 & 15.57 \\     
\midrule
\multirow{2}{*}{Mamba-based} & Wave-Mamba \cite{zou2024wave} & 22.76& 0.8419& \underline{0.0791}& \textbf{27.87}& \underline{0.8935}& \underline{0.0451} & 24.69 & 0.9271 & 0.0584 & 21.19 & 0.6391 & 0.1818 & 1.26 & 7.22 \\  
& RetinexMamba \cite{bai2024retinexmamba} & 23.15 &0.8210 &0.0876& 27.31& 0.8667& 0.0551 & \textbf{25.89} & \underline{0.9346} & 0.0389 & 20.88 & 0.6298 & \underline{0.1689} & 3.59 & 34.76 \\  
\midrule 
& CWNet & \underline{23.60} & \textbf{0.8496} & \textbf{0.0648} & \underline{27.39} & \textbf{0.9005} & \textbf{0.0383} & \underline{25.50} & \textbf{0.9362} & \textbf{0.0195} & \textbf{21.50} & \textbf{0.6397} & \textbf{0.1562} & 1.23 & 11.3 \\  
\bottomrule  
\end{tabular}  
}  
\caption{Quantitative comparison on LOL-v1~\cite{lol}, LOL-v2-Real~\cite{lol}, LOL-v2-Syn~\cite{lol}, and LSRW-Huawei~\cite{lsrw}. The \textbf{best} and \underline{second-best} results are shown in bold and underlined respectively. \textit{Please note that we did not use the GT-Mean strategy.}}  
\label{tab:com}  
\end{table*}

\subsection{Loss Function}

Our total loss consists of five parts: 
\begin{equation}
\resizebox{0.85\hsize}{!}{$
{\mathcal L}_{total}=\lambda_{1}{\mathcal L}_{2}+\lambda_{2}{\mathcal L}_{ssim} + \lambda_{3}{\mathcal L}_{per}  + \lambda_{4}{\mathcal L}_{ca}  + \lambda_{5}{\mathcal L}_{sem}$,}
\end{equation}
where $\lambda$ denotes the loss weights, we set $\lambda_{1},\lambda_{2},\lambda_{3},\lambda_{4},\lambda_{5}=[1.0, 0.3, 0.2, 0.01, 0.01]$. ${\mathcal L}_{2}$ represents the $l_{2}$ loss. ${\mathcal L}_{ssim}$ is the structure similarity loss. ${\mathcal L}_{per}$ is the perceptual loss, which constrains the features extracted from VGG to obtain better visual results. ${\mathcal L}_{2}$, ${\mathcal L}_{ssim}$ and ${\mathcal L}_{per}$ constrain the output $I_{pre}$ and ground truth $I_{gt}$ end-to-end.

\section{Experiments}
\label{experiments}

\begin{figure*}[t]
    \centering
    \includegraphics[width=1\linewidth]{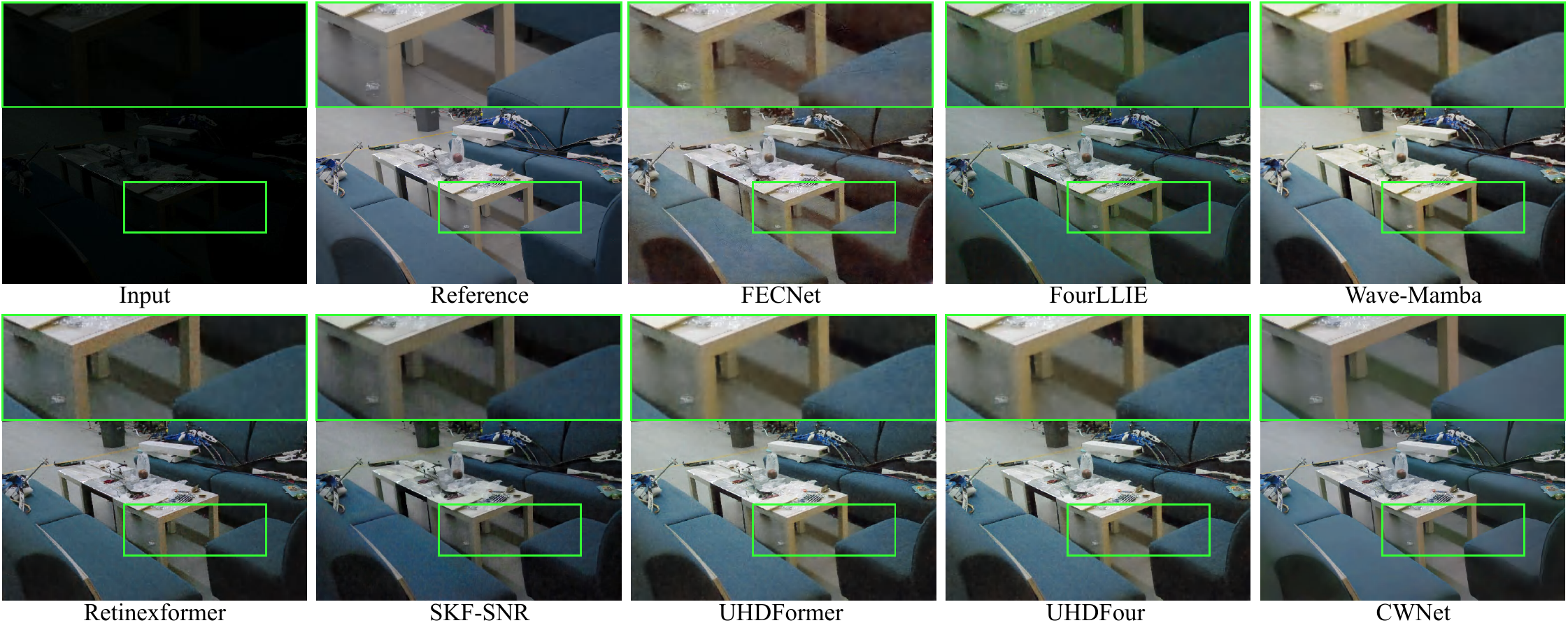}
    \caption{Visual comparison on LOL-v2-Real dataset.}
    \label{fig:lol-v2-com}
\end{figure*}

\begin{figure*}[t]
    \centering
    \includegraphics[width=1\linewidth]{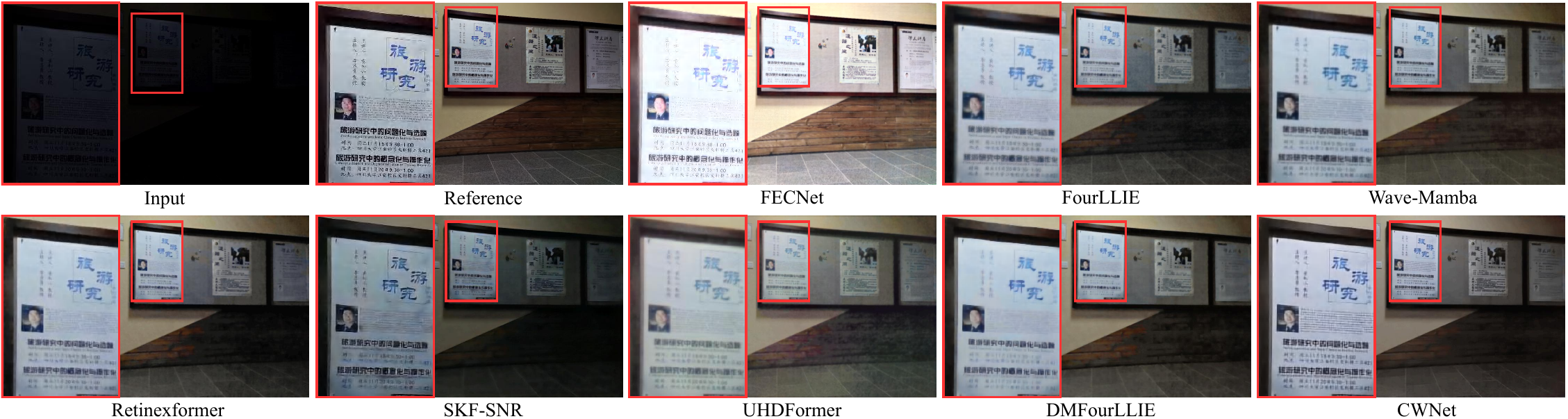}
    \caption{Visual comparison on LSRW-Huawei dataset.}
    \label{fig:huawei-com}
\end{figure*}

\subsection{Datasets and Experimental Setting}
CWNet is trained and evaluated on four LLIE datasets: LOL-v1~\cite{lol}, LOL-v2-Real~\cite{lol}, LOL-v2-Synthesis~\cite{lol}, and LSRW-Huawei~\cite{lsrw}. LOL-v1 contains 485 training and 15 testing pairs of real-world low-light/normal-light images. LOL-v2-Real provides 689 training and 100 testing pairs with more diverse real-world scenarios. For LOL-v2-Real evaluation, we use the model trained on LOL-v1 to demonstrate cross-dataset generalization. LOL-v2-Synthesis includes 900 training and 100 testing synthesized pairs. LSRW-Huawei comprises 2450 training and 30 testing pairs captured with different devices.

CWNet is implemented in PyTorch and trained end-to-end to jointly optimize all network parameters. The model employs a U-Net-like architecture with feature channels $16$ and asymmetric block configurations of $[1,3,4,3,1]$ and $[1,2,2,2,1]$ for low and high-frequency branches respectively. During training, input images are randomly cropped to $256 \times 256$ patches and augmented with random horizontal/vertical flips and rotations. We use the Adam optimizer with $\beta_1=0.9$, $\beta_2=0.99$, and an initial learning rate of $4.0 \times 10^{-4}$. The total training is conducted for $3.0 \times 10^{5}$ iterations with a batch size of 8.

\subsection{Comparison with Current Methods}
In this paper, our CWNet is compared to current state-of-the-art LLIE methods, including traditional approaches LIME~\cite{lime}, NPE~\cite{npe} and SRIE~\cite{retinex1}, deep learning-based methods Kind~\cite{kind}, Kind++~\cite{kind++}, MIRNet~\cite{lowlight9}, SGM~\cite{lol}, SNR-Aware~\cite{lowlight8}, FourLLIE~\cite{wang2023fourllie}, FECNet~\cite{huang2022deep}, UHDFour~\cite{li2023embedding}, Retinexformer~\cite{cai2023retinexformer}, DMFourLLIE~\cite{zhang2024dmfourllie}, RetinexMamba \cite{bai2024retinexmamba} and Wave-Mamba \cite{zou2024wave}.

\textbf{Quantitative Results on LOL-v2-Real, LOL-v2-Synthesis, and LSRW-Huawei Datasets.}
We comprehensively evaluate CWNet using PSNR, SSIM, and LPIPS metrics, where higher PSNR/SSIM and lower LPIPS indicate better image quality. As shown in Tab.~\ref{tab:com}, CWNet achieves superior performance across all benchmarks: PSNR of 23.60 dB on LOL-v1, 27.39 dB on LOL-v2-Real, 25.50 dB on LOL-v2-Synthesis, and 21.50 dB on LSRW-Huawei. Particularly noteworthy is the exceptional cross-dataset generalization from LOL-v1 training to LOL-v2-Real testing, achieving the best SSIM of 0.9005 and lowest LPIPS of 0.0383.
Importantly, CWNet achieves this superior performance while maintaining computational efficiency with only 1.23M parameters and 11.3G FLOPs, significantly outperforming parameter-heavy methods like MIRNet~\cite{lowlight9} (31.79M) and SNR-Aware~\cite{lowlight8} (39.12M), demonstrating effective balance between performance and efficiency.

\textbf{Visualization Comparison on LOL-v2-Real and LSRW-Huawei Datasets.}
We compare CWNet with state-of-the-art methods, including FECNet~\cite{huang2022deep}, FourLLIE~\cite{wang2023fourllie}, Wave-Mamba, Retinexformer~\cite{zou2024wave}, SKF-SNR~\cite{wu2023learning}, UHDFormer~\cite{wang2024uhdformer}, UHDFour~\cite{li2023embedding}, and DMFourLLIE~\cite{zhang2024dmfourllie}. As shown in Fig.~\ref{fig:lol-v2-com}, while other methods improve brightness, they often fail to maintain color and semantic consistency or control noise. For example, FECNet, FourLLIE, and Wave-Mamba show color deviations and noise, while Retinexformer and SKF-SNR lack sufficient brightening. UHDFormer and UHDFour perform better but still exhibit noise artifacts and lack smoothness. In contrast, CWNet produces clearer, more natural, and smoother results, ensuring semantic and color consistency, demonstrating its superior frequency domain modeling and structural fidelity.

\textbf{Visualization on LSRW-Huawei Dataset.}
As shown in Fig.~\ref{fig:huawei-com}, CWNet outperforms others in clarity and detail preservation. Methods like FECNet, SKF-SNR, and UHDFormer show exposure and brightening deficiencies, while CWNet achieves balanced brightness and excellent detail retention. For additional comparisons, refer to supplementary materials.

\subsection{Ablation Studies}

We conduct comprehensive ablation studies on the LSRW-Huawei dataset to evaluate the effectiveness of CWNet's design components. Tab.~\ref{tab:ablation} presents the results across two experimental settings:

\textbf{Component Removal Analysis.} We systematically remove key components to assess their individual contributions. Removing the causal inference mechanism results in a significant performance drop (PSNR: 20.87 dB vs. 21.53 dB). Similarly, ablating the feature extraction (FE) modules, HFEB, and LFEB leads to substantial performance degradation, with LFEB removal causing the most severe impact (PSNR drops to 20.41 dB), confirming the critical role of low-frequency processing in our dual-branch architecture.

\textbf{Component Replacement Analysis.} We validate the effectiveness of our proposed modules by replacing them with conventional alternatives. Substituting WTConv and FFTConv with standard convolutions reduces performance (PSNR: 21.42 dB and 21.36 dB respectively), highlighting the benefits of frequency-domain processing. Replacing HF-Mamba with standard VMamba's 2D-SSM structure~\cite{shi2024vmambair} also degrades performance (PSNR: 21.20 dB), demonstrating the superiority of our high-frequency Mamba design. Additionally, replacing semantic maps with global features shows performance reduction (PSNR: 21.48 dB), confirming the value of semantic guidance.

\textbf{Loss Weights Analysis.} We conducted comprehensive ablation experiments by systematically varying each weight parameter to analyze their individual impact on performance. As shown in Tab.\ref{tab:weight_ablation}, different weight configurations lead to varying performance across metrics. The baseline configuration achieves the best PSNR (21.53) and the second-best SSIM (0.6423), demonstrating strong perceptual quality.

These ablation results collectively validate that each proposed component contributes meaningfully to CWNet's superior performance. For additional ablation studies and downstream applications, please refer to our supplementary materials.

\begin{table}[t]  
\centering  
\renewcommand{\arraystretch}{1.2} 
\setlength{\tabcolsep}{8pt} 
\resizebox{0.9\linewidth}{!}{  
\begin{tabular}{@{}l|c|c|c@{}}  
\toprule  
Ablation Settings & PSNR↑ & SSIM↑ & LPIPS↓ \\ 
\midrule   
\multicolumn{4}{c}{Component Removal} \\
\midrule
w/o Casual Inference  & 20.87 & 0.6375 & 0.1781 \\   
w/o FE & 20.98 & 0.6387 & 0.1804 \\   
w/o HFEB & 20.58 & 0.6317 & 0.1903 \\   
w/o LFEB & 20.41 & 0.6302 & 0.1985 \\   
\midrule
\multicolumn{4}{c}{Component Replacement} \\
\midrule
WTConv $\rightarrow$ Conv & 21.42 & 0.6415 & 0.1690 \\   
FFTConv $\rightarrow$ Conv & 21.36 & 0.6396 & 0.1721 \\   
HF-Mamba $\rightarrow$ VMamba (2D-SSM) & 21.20 & 0.6394 & 0.1735 \\   
Segmantic Maps $\rightarrow$ Global & 21.48 & 0.6417 & 0.1652 \\   
\midrule
CWNet & \textbf{21.53} & \textbf{0.6423} & \textbf{0.1631} \\   
\bottomrule  
\end{tabular}}  
\vspace{-0.2cm}
\caption{Ablation experiment study of CWNet. By designing ablation experiments on component removal and replacement, the effectiveness of each component and composition of CWNet is fully verified.}  
\label{tab:ablation}  
\end{table}

\begin{table}[t]  
\centering  
\renewcommand{\arraystretch}{1.2} 
\setlength{\tabcolsep}{6pt} 
\resizebox{0.98\linewidth}{!}{ 
\begin{tabular}{@{}c|c|c|c|c|c|c|c|c@{}}  
\toprule  
\textbf{Loss} & ${\mathcal L}_{1}$ & ${\mathcal L}_{ssim}$ & ${\mathcal L}_{per}$ & ${\mathcal L}_{ca}$ & ${\mathcal L}_{sem}$ & \textbf{PSNR↑} & \textbf{SSIM↑} & \textbf{LPIPS↓} \\
\midrule  
 & 1.0 & 0.3 & 0.2 & 0.01 & 0.01 & \textbf{21.53} & \underline{0.6423} & 0.1631 \\
A & 1.0 & 0.4 & 0.2 & 0.01 & 0.01 & 21.39 & \textbf{0.6433}& \textbf{0.1597} \\
B & 1.0 & 0.3& 0.3 & 0.01 & 0.01 & 21.34 & 0.6407 & \underline{0.1601} \\
C & 1.0 & 0.3 & 0.2 & 0.05 & 0.01 & \underline{21.43} & 0.6386 & 0.1614 \\
D & 1.0 & 0.3 & 0.2 & 0.001 & 0.01 & 21.17 & 0.6408 & 0.1651 \\
E & 1.0 & 0.3 & 0.2 & 0.01 & 0.05 & 20.89 & 0.6382 & 0.1701 \\
F & 1.0 & 0.3 & 0.2 & 0.01 & 0.001 & 20.94 & 0.6371 & 0.1652 \\
\bottomrule  
\end{tabular}}  
\caption{Ablation Study on Weight Configuration in Loss Functions. The loss function is defined as ${\mathcal L}_{total}=\lambda_{1}{\mathcal L}_{2}+\lambda_{2}{\mathcal L}_{ssim} + \lambda_{3}{\mathcal L}_{per}  + \lambda_{4}{\mathcal L}_{ca}  + \lambda_{5}{\mathcal L}_{sem}$. The \textbf{best} and \underline{second-best} results are shown in bold and underlined respectively.}  
\vspace{-0.4cm}
\label{tab:weight_ablation}  
\end{table}

\subsection{Limitations}

\begin{figure}[t]
    \centering
    \includegraphics[width=1\linewidth]{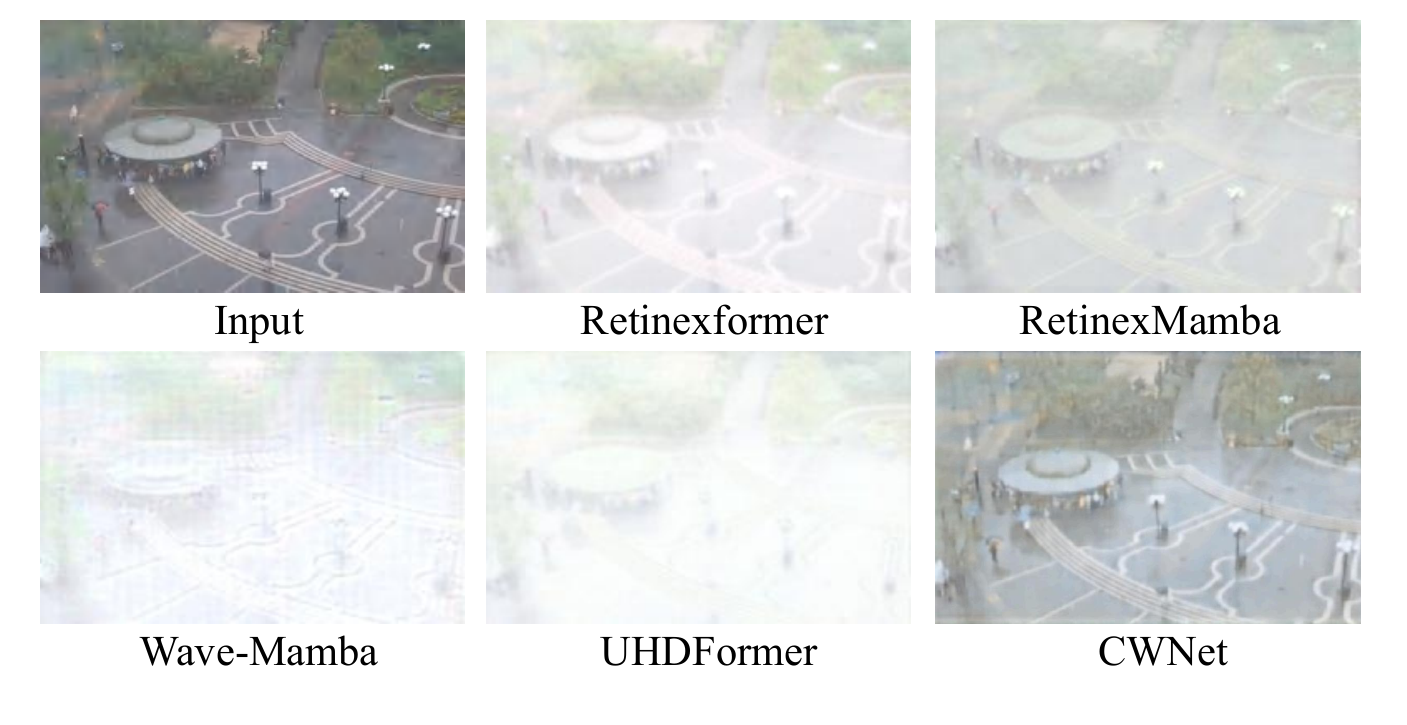}
    \caption{Failure cases in multiple degradation scenarios.}
      \vspace{-0.4cm}
    \label{fig:fail_case}
\end{figure}

As shown in Fig.\ref{fig:fail_case}, when facing its own degradation, such as blurring or haze, CWNet maintains the color and lighting intensity of the image as much as possible compared to other methods, yet the restoration quality is subpar. This opens up new avenues for us to explore how to ensure more effective recovery of low-light images that experience multiple degradation conditions simultaneously.

\begin{figure}[t]
  \centering
  \includegraphics[width=0.48\textwidth]{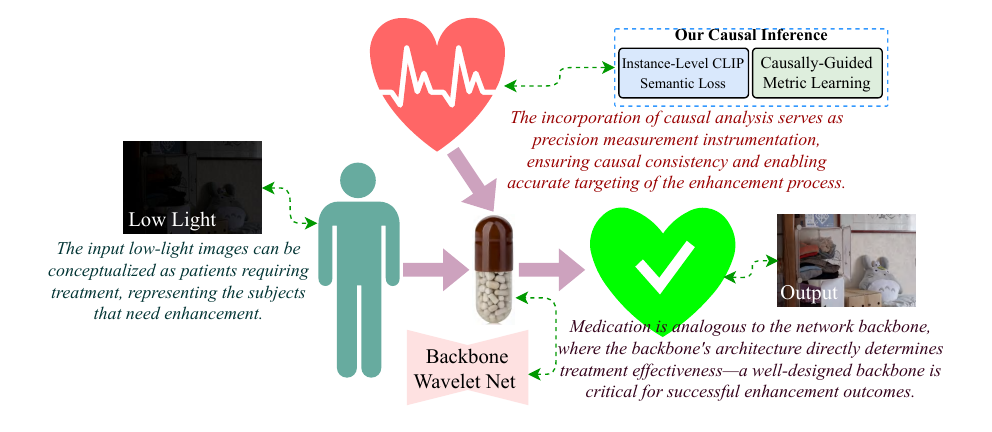}
  \caption{The connection between wavelet and causality.}
  \label{fig:Causal-and-Wavelet}
\end{figure}

\section{Discussion and Conclusion}
\label{conlusion}

\textbf{Causality-Wavelet Connection.}
The interpretability of CWNet and causality-wavelet connection:
Wavelet structure and causal analysis are organically integrated (Fig.\ref{fig:Causal-and-Wavelet}) with causal treatment providing model interpretability. We clarify: \textbf{a)} Causal Perspective: Our focus centers on causal analysis (analogous to measurement instruments in causal treatment), treating scene information as causal factors in LLIE, ensuring causal consistency during enhancement.
\noindent \textbf{b)} Wavelet-based Backbone: To achieve causal factor consistency (analogous to medication in treatment), our design leverages low-frequency enhancement for color and brightness consistency while high-frequency components incorporate Mamba consistency scanning to enhance detail modeling and promote structural consistency.

\textbf{Conclusion.}
In this paper, we proposed the Causal Wavelet Network (CWNet), a novel architecture that integrates causal inference with wavelet transform to address low-light image enhancement. By leveraging causal analysis, we effectively separated causal factors from non-causal interference, ensuring both global and local semantic consistency. The HFRB further refined feature extraction by modeling wavelet frequency domain characteristics. Extensive experiments demonstrate that CWNet achieves superior performance over state-of-the-art methods, highlighting the effectiveness of causal reasoning in enhancing image quality. This work underscores the potential of causal inference as a powerful tool for advancing low-light image enhancement.

\section*{Acknowledgements}
This work was supported by Jilin Province Industrial Key Core Technology Tackling Project (20230201085GX).

{
    \small
    \bibliographystyle{ieeenat_fullname}
    \bibliography{main}
}

\setcounter{section}{0}
\renewcommand{\thesection}{\Alph{section}}  

\clearpage
\setcounter{page}{1}
\maketitlesupplementary

\section{Network Structure}  
\label{sec:network}  
\subsection{Detailed Structure of Feature Extraction (FE)}  
In the FE module, we utilize WTConv \cite{finder2025wavelet}, H-Conv, V-Conv, and D-Conv. Each component is described in detail below:  

WTConv \cite{finder2025wavelet} employs a convolution kernel of size 5 and is configured with 3 levels of wavelet downsampling to progressively reduce spatial resolution while preserving frequency information.   

The \textbf{H-Conv, V-Conv, and D-Conv} modules utilize specifically designed convolution kernels to capture directional features in the input data. These kernels are as follows:  

\begin{itemize}  
    \item Horizontal Convolution (H-Conv):  
    \begin{align*}  
    \text{Horizontal Kernel:} \quad   
    \begin{bmatrix}   
    1 & 0 & -1 \\   
    1 & 0 & -1 \\   
    1 & 0 & -1   
    \end{bmatrix}  
    \end{align*}  
    This kernel is designed to emphasize horizontal edges in the input data.  

    \item Vertical Convolution (V-Conv):  
    \begin{align*}  
    \text{Vertical Kernel:} \quad   
    \begin{bmatrix}   
    1 & 1 & 1 \\   
    0 & 0 & 0 \\   
    -1 & -1 & -1   
    \end{bmatrix}  
    \end{align*}  
    This kernel is designed to capture vertical edge features.  

    \item Diagonal Convolution (D-Conv):  
    \begin{align*}  
    \text{Diagonal Kernel:} \quad   
    \begin{bmatrix}   
    0 & 1 & 0 \\   
    1 & -4 & 1 \\   
    0 & 1 & 0   
    \end{bmatrix}  
    \end{align*}  
    This kernel is designed to detect diagonal structures.  
\end{itemize}

\begin{figure}[t] 
    \centering
    \includegraphics[width=1.0\linewidth]{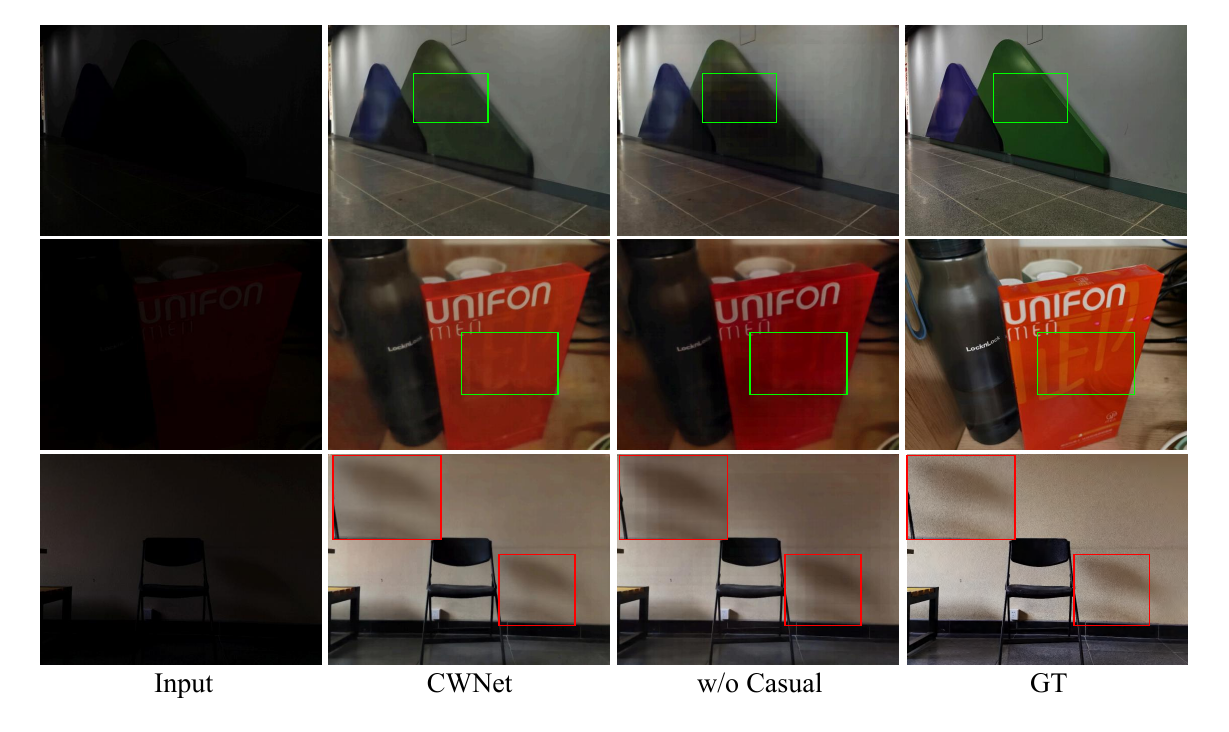}
    \caption{Visual comparison of results with and without the causal inference. Focus on the highlighted regions to observe differences in color and brightness, demonstrating the module's effectiveness.}
    \label{fig:ab-cas}
\end{figure}

\section{Experiments}  
\subsection{Visualization Comparison on LOL-v2-Synthesis Dataset.}  
\begin{figure*}[t]
    \centering
    \includegraphics[width=1\linewidth]{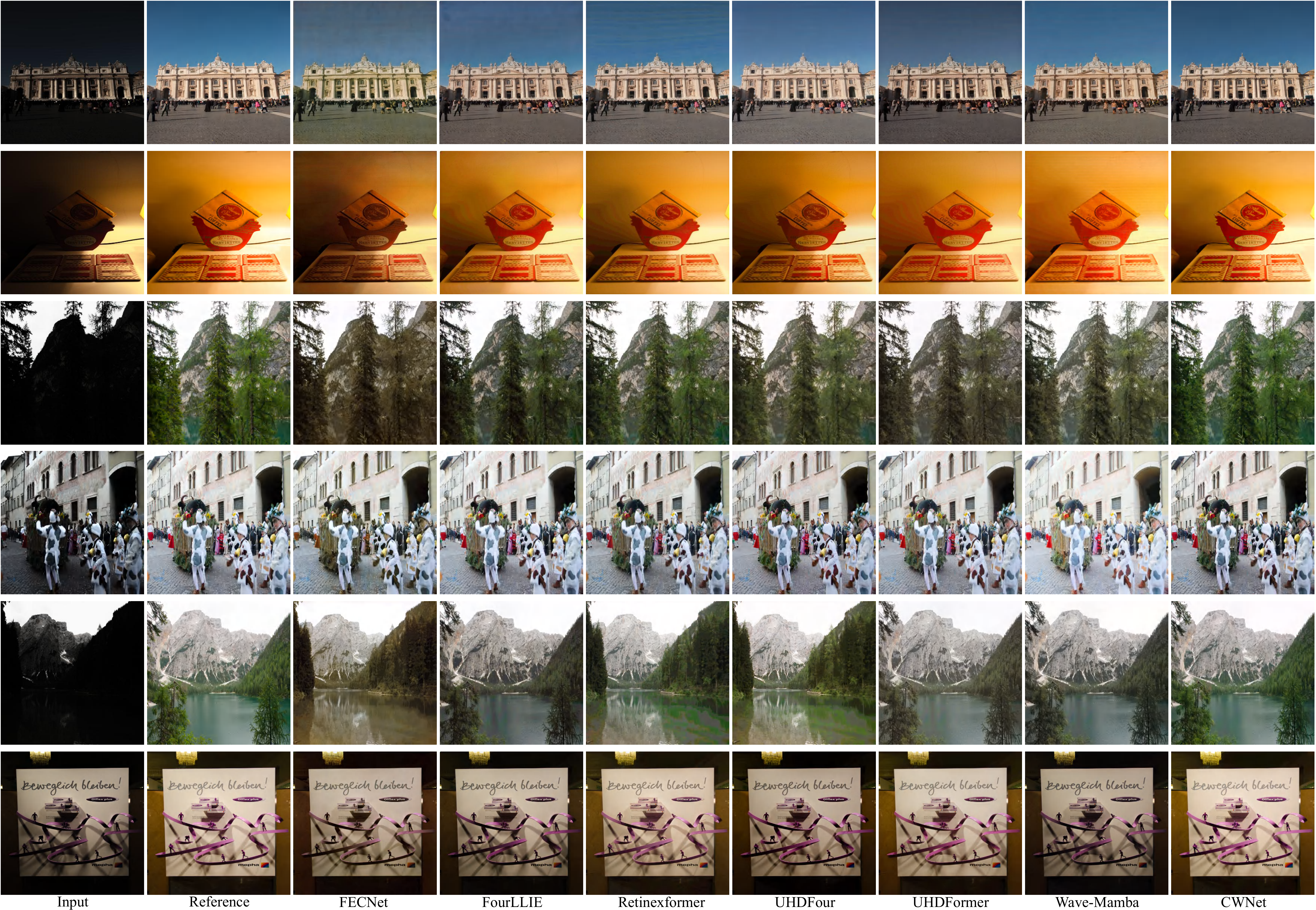}
    \caption{Visual comparison on LOL-v2-Synthesis dataset.}
    \label{fig:com-syn}
\end{figure*}

Fig.\ref{fig:com-syn} presents a comparison on the LOL-v2-Synthesis dataset between our CWNet and current state-of-the-art methods, including FECNet \cite{huang2022deep}, FourLLIE \cite{wang2023fourllie}, Retinexformer \cite{cai2023retinexformer}, UHDFour \cite{li2023embedding}, UHDFormer \cite{wang2024uhdformer}, and Wave-Mamba \cite{zou2024wave}.  

In the first row, other methods exhibit stripe noise in the sky region. Specifically, FECNet shows severe hue distortion, while other methods lack image sharpness. In the second row, similar artifacts are observed, with stripe-like noise appearing in the background areas.   

In the third and fifth rows, other methods exhibit color distortion when compared to the reference ground truth. In contrast, our results appear more natural and exhibit better visual quality. This demonstrates the effectiveness of our causal inference component in mitigating color distortion and preserving semantic structures by disentangling causal relationships in the feature space.   

In the fourth and last rows, upon magnification, our results exhibit the most consistent and visually pleasing brightening effect. The comparison on the LOL-v2-Synthesis dataset further demonstrates that our CWNet achieves natural brightness restoration and excels in preserving semantic and color consistency.

\begin{table}[t]  
\centering  
\renewcommand{\arraystretch}{1.2} 
\setlength{\tabcolsep}{6pt} 
\resizebox{1\linewidth}{!}{%
\begin{tabular}{l|c|c|c|c|c|c}  
\hline  
Methods & DICM & LIME & MEF & NPE & VV & AVG \\   
\hline  
Kind~\cite{kind} & \textbf{3.61} & 4.77 & 4.82 & 4.18 & 3.84 & 4.24 \\   
MIRNet~\cite{lowlight9} & 4.04 & 6.45 & 5.50 & 5.24 & 4.74 & 5.19 \\   
SGM \cite{lol} & 4.73 & 5.45 & 5.75 & 5.21 & 4.88 & 5.21 \\   
FECNet \cite{huang2022deep} & 4.14 & 6.04 & 4.71 & 4.50 & 3.75 & 4.55 \\   
HDMNet~\cite{liang2022learning} & 4.77 & 6.40 & 5.99 & 5.11 & 4.46 & 5.35 \\   
Bread~\cite{guo2023low} & 4.18 & 4.72 & 5.37 & 4.16 & \textbf{3.30} & 4.35 \\   
Retinexformer \cite{cai2023retinexformer} & 4.01 & \textbf{3.44} & \underline{3.73} & \underline{3.89} & \underline{3.71} & \underline{3.76} \\   
UHDFormer \cite{wang2024uhdformer} & 4.42 & 4.35 & 4.74 & 4.40 & 4.28 & 4.44 \\
Wave-Mamba \cite{zou2024wave} & 4.56 & 4.45 & 4.76 & 4.54 & 4.71 & 4.60 \\    
\hline  
CWNet & \underline{3.92} & \underline{3.58} & \textbf{3.66} & \textbf{3.61} & 3.74 & \textbf{3.70} \\    
\hline  
\end{tabular}}  
\caption{NIQE scores on DICM, LIME, MEF, NPE, and VV datasets. Lower NIQE scores indicate better perceptual quality. The \textbf{best} and \underline{second-best} results in each column are shown in bold and underlined respectively. "AVG" denotes the average NIQE scores across the five datasets.}  
\label{tab:comsmall}  
\end{table}

\subsection{Quantitative and Qualitative Comparison on the DICM, LIME, MEF, NPE, and VV Datasets}  
\textbf{Quantitative Comparison.}  
We evaluated CWNet on five independent datasets: DICM~\cite{dicm} (69 images), LIME~\cite{lime} (10 images), MEF~\cite{mefl} (17 images), NPE~\cite{npe} (85 images), and VV (24 images). The evaluation was conducted using the no-reference image quality assessment metric NIQE~\cite{6353522}, with lower scores indicating better perceptual quality and naturalness. Tab.~\ref{tab:comsmall} presents the NIQE results. As shown, CWNet outperforms several state-of-the-art (SOTA) methods, including SKF-SNR, UHDFormer, and Wave-Mamba. To ensure a fair comparison, all methods were pretrained on the LSRW-Huawei dataset.

\textbf{Qualitative Comparison.}  
We provide qualitative comparisons on the DICM, LIME, MEF, NPE, and VV datasets to visually demonstrate the effectiveness of CWNet.

\begin{figure*}[t]
    \centering
    \includegraphics[width=1\linewidth]{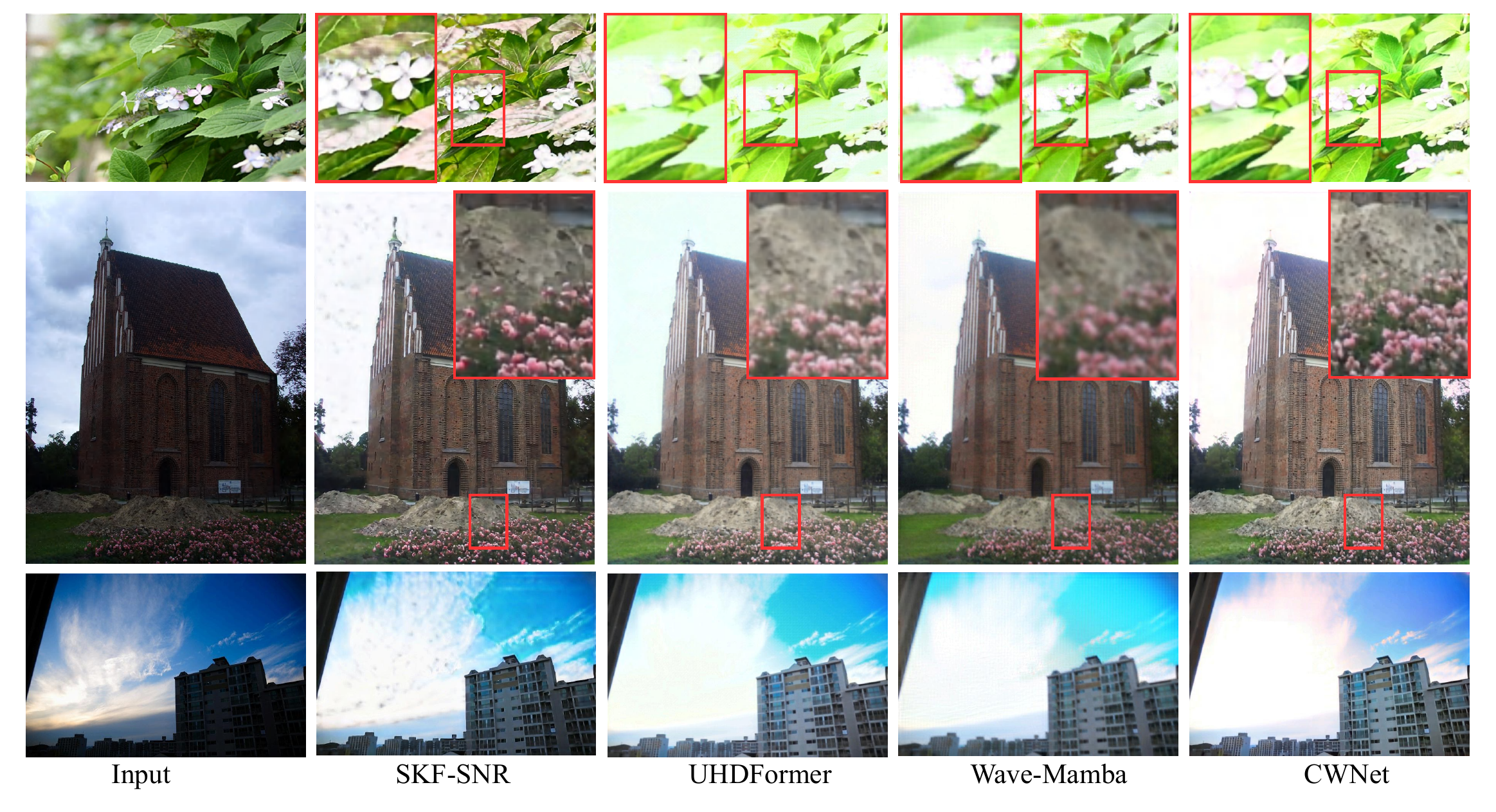}
    \caption{Visual comparison on DICM dataset.}
    \vspace{-0.5cm}
    \label{fig:DICM}
\end{figure*}

1) Fig.\ref{fig:DICM} shows the qualitative comparison on the DICM dataset. In the first row, SKF-SNR produces distorted and overexposed results, which negatively impact the visual quality. UHDFormer generates relatively natural results but suffers from overexposure, particularly in the highlighted regions, where the flower colors are overly brightened. Wave-Mamba exhibits blurred edge details and insufficient exposure control. In contrast, CWNet produces clearer and more natural enhancement results. In the second row, CWNet produces sharper and more natural results, while SKF-SNR suffers from underexposure and noise artifacts. UHDFormer and Wave-Mamba produce relatively blurry results. In the third row, SKF-SNR introduces unavoidable noise, and both UHDFormer and Wave-Mamba generate unnatural sky colors. CWNet, however, produces results with consistent and natural sky colors, demonstrating its robustness in handling color distortions.  

\begin{figure*}[t]
    \centering
    \includegraphics[width=1\linewidth]{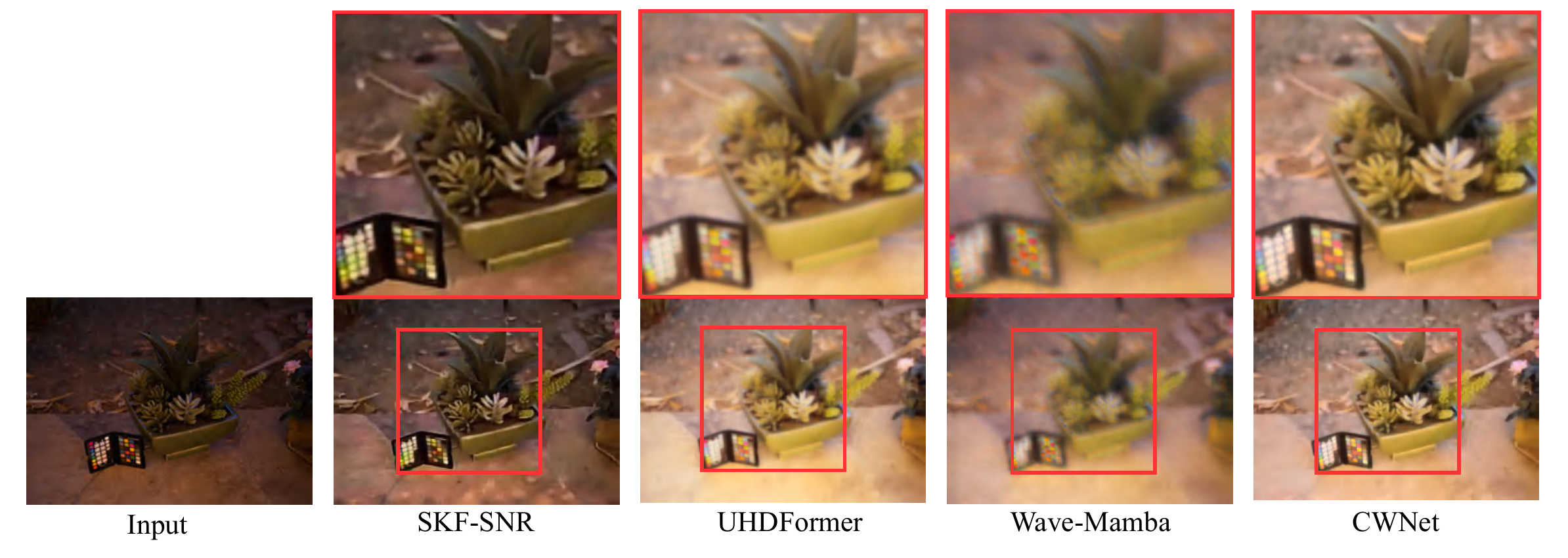}
    \caption{Visual comparison on LIME dataset.}
    \label{fig:LIME}
\end{figure*}

2) Fig.\ref{fig:LIME} presents the qualitative comparison on the LIME dataset. SKF-SNR fails to achieve sufficient brightness enhancement, while UHDFormer and Wave-Mamba produce reasonable brightness but suffer from blurriness and lack of detail. In contrast, CWNet generates sharper textures and richer details, further validating the robustness of our high- and low-frequency modeling in capturing fine-grained details and global structures.

\begin{figure*}[t]
    \centering
    \includegraphics[width=1\linewidth]{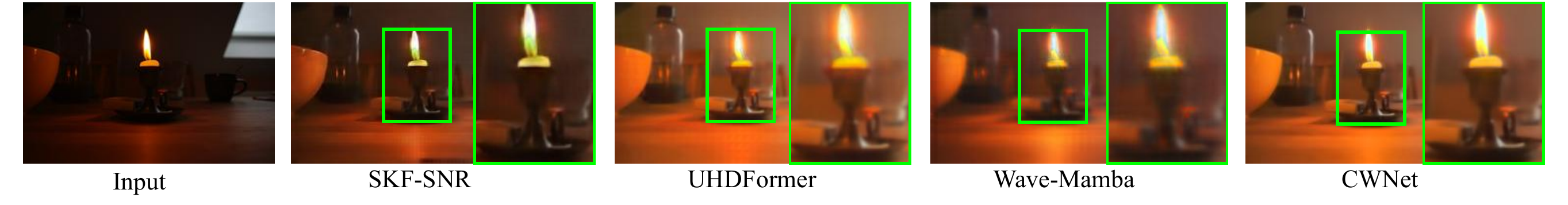}
    \caption{Visual comparison on MEF dataset.}
    \label{fig:MEF}
\end{figure*}

3) Fig.\ref{fig:MEF} illustrates the qualitative comparison on the MEF dataset. SKF-SNR produces unnatural flame colors and insufficient brightness enhancement. UHDFormer and Wave-Mamba exhibit varying degrees of blurriness, while CWNet achieves the clearest and most visually pleasing enhancement results.

\begin{figure*}[t] 
    \centering
    \includegraphics[width=1\linewidth]{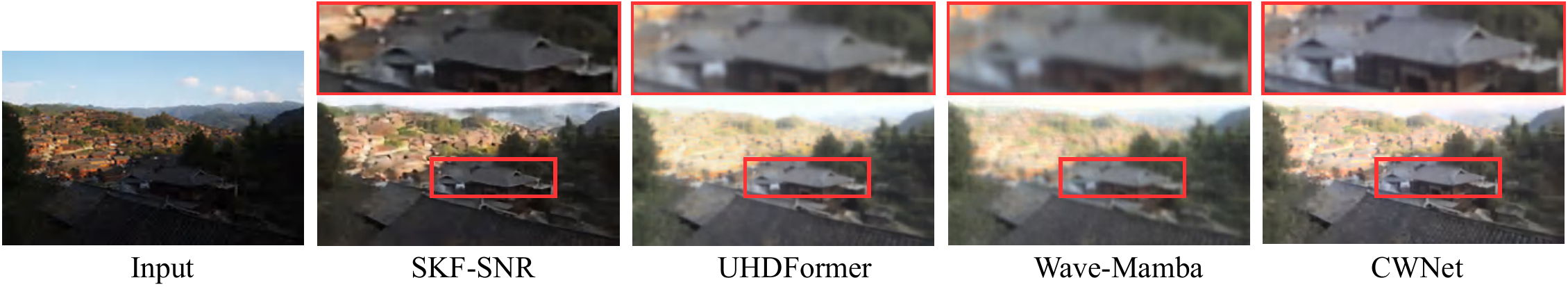}
    \caption{Visual comparison on NPE dataset.}
    \label{fig:NPE}
\end{figure*}

4) Fig.\ref{fig:NPE} shows the qualitative comparison on the NPE dataset. Similar to the MEF dataset, SKF-SNR fails to provide sufficient brightness enhancement, and both UHDFormer and Wave-Mamba suffer from blurriness. CWNet, on the other hand, produces visually superior results with richer detail information.

\begin{figure*}[t] 
    \centering
    \includegraphics[width=1\linewidth]{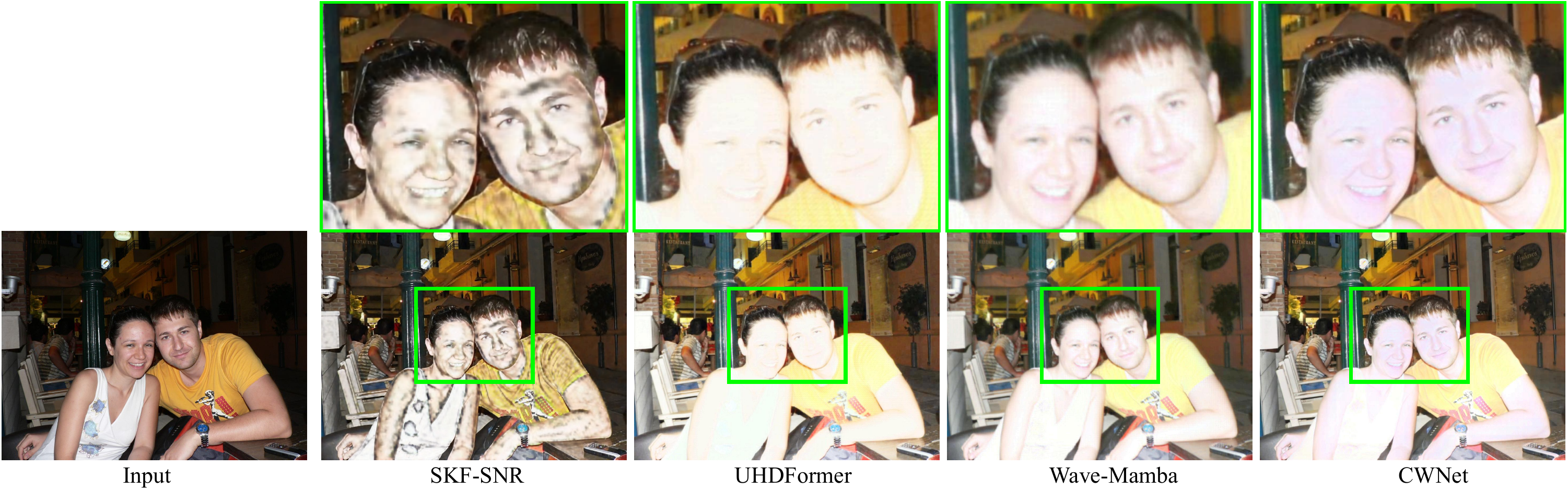}
    \caption{Visual comparison on VV dataset.}
    \label{fig:VV}
\end{figure*}

5) Fig.\ref{fig:VV} demonstrates the qualitative comparison on the VV dataset. SKF-SNR generates distorted enhancement results with significant noise artifacts. UHDFormer suffers from overexposure, as observed in the highlighted facial regions, while Wave-Mamba produces blurry results. In contrast, CWNet produces natural enhancements with well-preserved details.

\begin{table}[t] 
\centering  
\resizebox{0.95\linewidth}{!}{  
\begin{tabular}{@{}lcccc@{}}  
\toprule  
Metrics & UHDFour \cite{li2023embedding} & UHDFormer \cite{wang2024uhdformer} & Wave-Mamba \cite{zou2024wave} & CWNet \\ \midrule  
UICM ↑  & 0.7464   & 0.9571  & 0.9082 & \textbf{0.9663} \\   
NIQE ↓  & 5.385    & 5.564   &  5.689 & \textbf{4.494}  \\   
\bottomrule  
\end{tabular}}  
\caption{Visualization quality comparison on DarkFace.  The \textbf{best} results in each column are shown in bold.}  
\label{tab:uicm-niqe}  
\end{table}

\subsection{Quantitative and Qualitative Comparison on the DarkFace Dataset}  
We conducted quantitative and qualitative comparisons against current state-of-the-art (SOTA) methods on the DarkFace dataset~\cite{yang2020advancing}. The DarkFace dataset, with 6,000 real-world low-light images, serves as a challenging benchmark for low-light image enhancement.  

For quantitative evaluation, we employed two metrics: NIQE and the Underwater Image Colorfulness Measure (UICM)\cite{schechner2005recovery}, which is commonly used to evaluate colorfulness and naturalness in enhanced images. Unlike NIQE, where lower scores indicate better perceptual quality, higher UICM scores reflect greater colorfulness and naturalness. As shown in Tab.\ref{tab:uicm-niqe}, our method achieves the best performance across both metrics, highlighting its effectiveness in low-light image enhancement.

\begin{figure*}[t]
    \centering
    \includegraphics[width=1\linewidth]{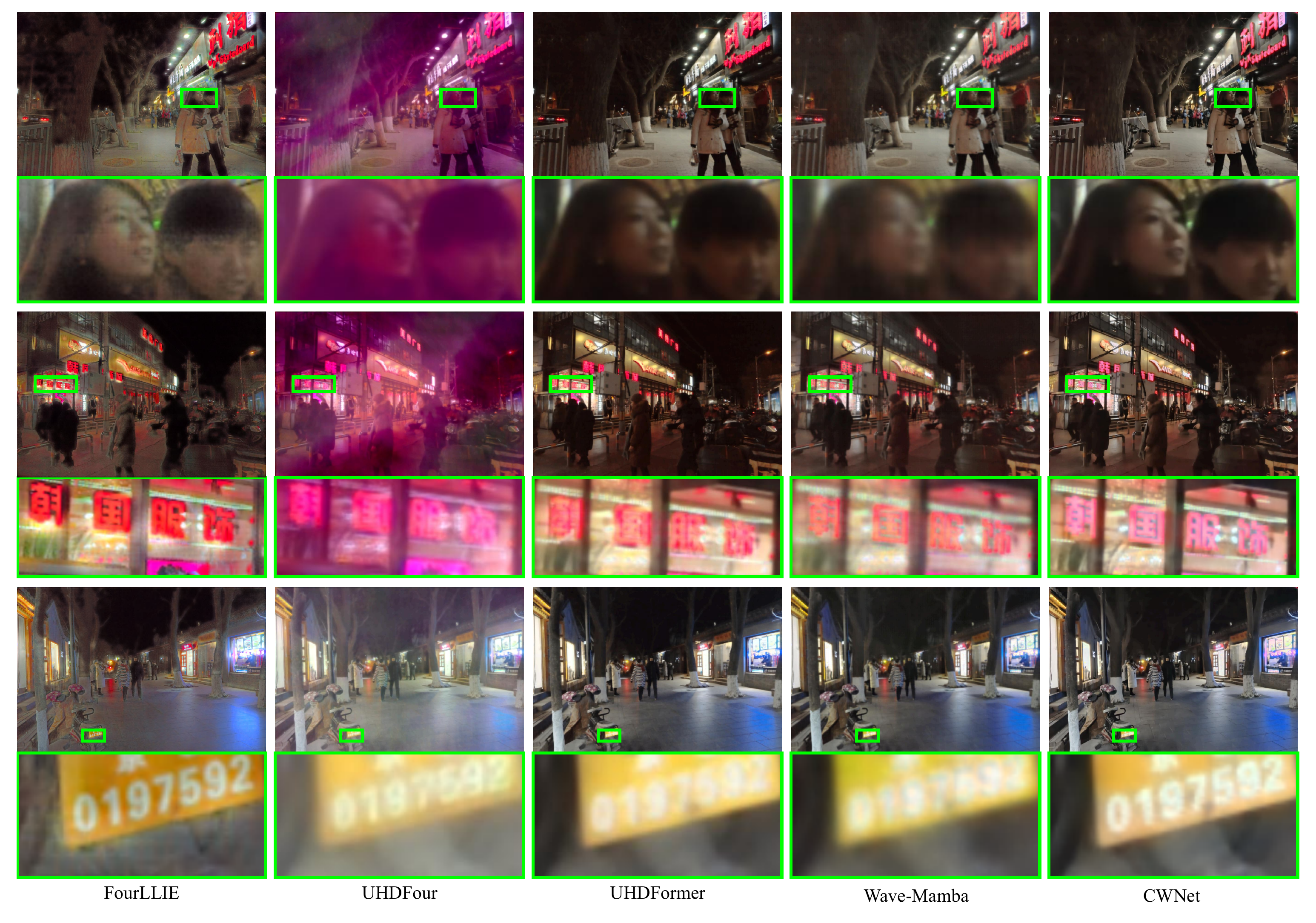}
    \caption{Visual comparison on DarkFace dataset.}
    \label{fig:DarkFace}
\end{figure*}

Fig.\ref{fig:DarkFace} illustrates the qualitative comparison on the DarkFace dataset. UHDFormer produces severe color distortions, particularly in bright regions, indicating its limited generalization ability on this dataset. In the first and last rows, the zoomed-in regions show that our method outperforms UHDFormer and Wave-Mamba in detail preservation and sharpness. Furthermore, in the second row, our method demonstrates superior exposure control, yielding natural and visually pleasing results. These observations validate the effectiveness of CWNet in low-frequency brightness control for natural exposure and high-frequency detail enhancement for sharper textures.

\section{Ablation Study and Downstream Application}

\subsection{Ablation Study: Validating the Effectiveness of Core Modules}
Fig.\ref{fig:ab-cas} demonstrates the impact of causal inference on CWNet's performance. By focusing on the highlighted regions, it is evident that removing causal inference leads to inconsistencies in brightness and semantic coherence. In contrast, incorporating causal inference ensures both brightness restoration and semantic consistency, effectively preserving color fidelity and structural details. This validates the module's ability to model causal relationships, enhancing overall enhancement quality.

\subsection{Ablation Study on Negative Sample Quantity in the Causal Inference}
We conducted an ablation study within the Causal Inference module to determine the optimal number of negative samples for brightness degradation $I_{l}$ (simulating underexposed regions) and color anomaly $I_{c}$ (introducing color distortions), as summarized in Tab.\ref{tab:negative_sample_ablation}. The results indicate that the optimal performance is achieved when the number of samples for both brightness degradation and color anomaly is set to 3.  

\begin{table}[t]  
\centering  
\renewcommand{\arraystretch}{1.2} 
\setlength{\tabcolsep}{10pt} 
\resizebox{0.6\linewidth}{!}{ 
\begin{tabular}{@{}lccc@{}}  
\toprule  
\textbf{Number} & \textbf{PSNR↑} & \textbf{SSIM↑} & \textbf{LPIPS↓} \\
\midrule  
L=1, C=1 & 21.06 & 0.6381 & 0.1772 \\
L=2, C=2 & \textbf{21.50} & 0.6397 & \textbf{0.1562} \\
L=3, C=3 & 21.34 & \textbf{0.6401} & 0.1587 \\
\bottomrule  
\end{tabular}}  
\caption{Ablation Study on Negative Sample Quantity in the Causal Inference Module. The first column represents the number of $I_{l}$ and $I_{c}$. The \textbf{best} results in each column are shown in bold.}   
\label{tab:negative_sample_ablation}  
\end{table}

\subsection{Object Detection on DarkFace Dataset}  
\begin{figure*}[t] 
    \centering
    \includegraphics[width=1\linewidth]{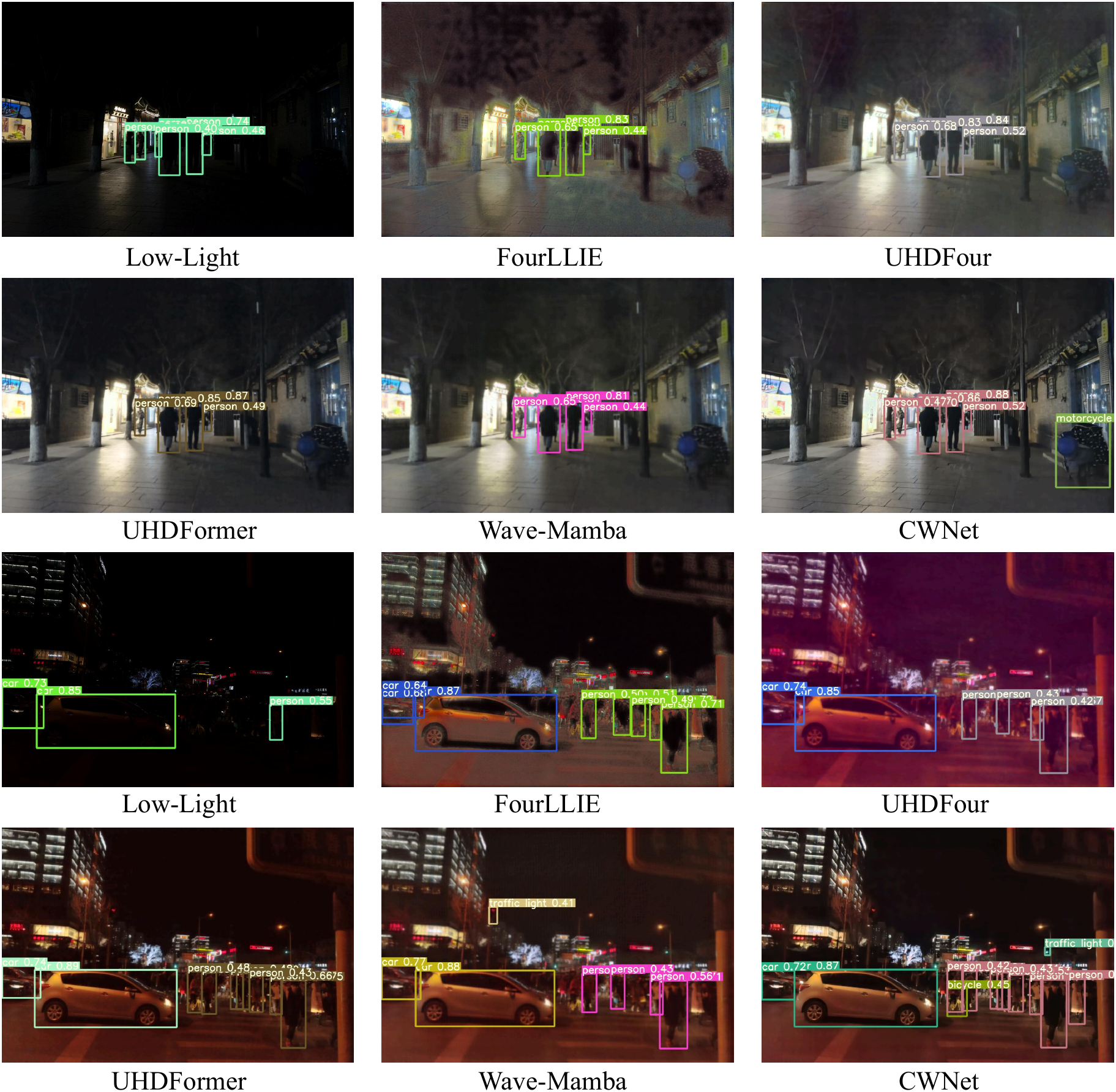}
    \caption{Detection comparison results on DarkFace dataset.}
    \label{fig:detection}
\end{figure*}

To evaluate how enhanced images affect downstream tasks, we conducted object detection experiments on the DarkFace dataset~\cite{yang2020advancing}. We evaluated our method on 200 randomly selected images from the dataset using the official YOLOv5 model pretrained on the COCO dataset~\cite{lin2014microsoft}, as YOLOv5 is a widely used object detection framework and COCO provides a diverse set of object categories. Fig.\ref{fig:detection} presents the visual comparison results, showing that our method achieves superior detection accuracy compared to others.  

In the first experiment, our method achieves higher confidence scores for detected pedestrians compared to other methods and uniquely detects the motorcycle on the right, likely due to its ability to enhance fine-grained details in low-light conditions. In the second experiment, our method detects more pedestrians with higher confidence and correctly identifies the traffic light, a task that is particularly challenging due to the small size and low contrast of the traffic light in the original image. In contrast, Wave-Mamba misclassifies building lights as traffic lights.  

These results demonstrate that our enhancement method generates clearer images while preserving semantic structures effectively. By leveraging causal inference, CWNet emphasizes intrinsic image content, enhancing downstream task performance such as object detection.

\subsection{Superior Edge Detection Performance}
\begin{figure*}[t] 
    \centering
    \includegraphics[width=1\linewidth]{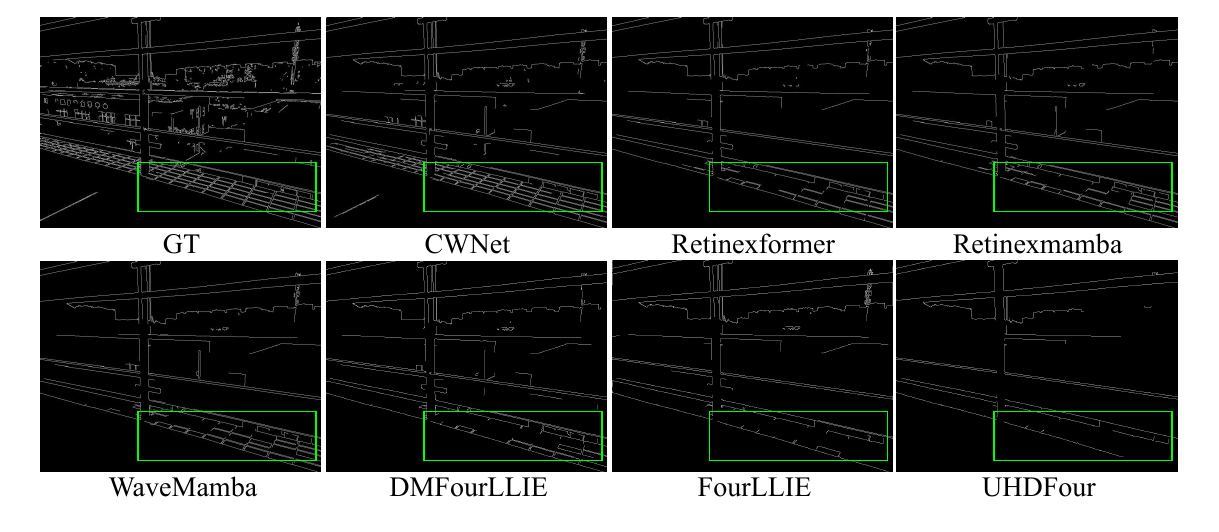}
    \caption{Edge detection comparison shows our method restores details more precisely, especially in highlighted regions, validating the high-frequency module.}
    \label{fig:canny}
\end{figure*}

Fig.~\ref{fig:canny} compares edge detection performance across state-of-the-art methods, highlighting CWNet's ability to restore details more precisely, particularly in highlighted regions. This superior performance validates the effectiveness of CWNet's high-frequency module in preserving fine details and structural fidelity, further demonstrating the robustness of our architecture.

\end{document}